\documentclass[journal]{IEEEtran}

\usepackage{graphicx}
\usepackage[numbers]{natbib}
\usepackage{amssymb}
\usepackage{amsmath}
\usepackage{centernot}
\usepackage{amsfonts}
\usepackage{algorithmic,algorithm}
\usepackage{color}
\usepackage{soul}
\usepackage{adjustbox}
\usepackage{subfig}
\usepackage{url}
\usepackage{pifont}

\begin{document}
%
\title{Deep Hierarchical Reinforcement Learning Algorithm in Partially Observable Markov Decision Processes$^{\textbf{\ding{78}}}$}
%
%
%
\author{\IEEEauthorblockN{Le Pham Tuyen\IEEEauthorrefmark{1}, Ngo Anh Vien\IEEEauthorrefmark{2}, Md. Abu Layek\IEEEauthorrefmark{1}, TaeChoong Chung \IEEEauthorrefmark{1} $^{\textbf{\ding{53}}}$} \\
    \IEEEauthorblockA{\IEEEauthorrefmark{1}Artificial Intelligence Lab, Computer Science and Engineering Department, Kyung Hee University,
Yongin, Gyeonggi 446-701, South Korea \\
    \{tuyenple, layek, tcchung\}@khu.ac.kr} \\ 
    \IEEEauthorblockA{\IEEEauthorrefmark{2}EEECS/ECIT, Queen’s University Belfast, Belfast, UK \\
    v.ngo@qub.ac.uk}

\thanks{\ding{53} Corresponding author}
\thanks{\ding{78} This work has been submitted to the IEEE for possible publication. Copyright may be transferred without notice, after which this version may no longer be accessible.}}

%
%

\maketitle

\begin{abstract}
In recent years, reinforcement learning has achieved many remarkable successes due to the growing adoption of deep learning techniques and the rapid growth in computing power. Nevertheless, it is well-known that flat reinforcement learning algorithms are often not able to learn well and data-efficient in tasks having hierarchical structures, e.g. consisting of multiple subtasks. Hierarchical reinforcement learning is a principled approach that is able to tackle these challenging tasks. On the other hand, many real-world tasks usually have only partial observability in which state measurements are often imperfect and partially observable. The problems of RL in such settings can be formulated as a partially observable Markov decision process (POMDP). In this paper, we study hierarchical RL in POMDP in which the tasks have only partial observability and possess hierarchical properties. We propose a hierarchical deep reinforcement learning approach for learning in hierarchical POMDP. The deep hierarchical RL algorithm is proposed to apply to both MDP and POMDP learning. We evaluate the proposed algorithm on various challenging hierarchical POMDP.
\end{abstract}

\begin{IEEEkeywords}
Hierarchical Deep Reinforcement Learning, Partially Observable MDP (POMDP), Semi-MDP, Partially Observable Semi-MDP (POSMDP)
\end{IEEEkeywords}

%
\IEEEpeerreviewmaketitle

\section{Introduction} 
\label{intro}
 
Reinforcement Learning (RL) \citep{Sutton:1998:IRL:551283} is a subfield of machine learning focused on learning a policy in order to maximize total cumulative reward in an unknown environment. RL has been applied to many areas such as robotics \citep{deisenroth2013survey}\citep{peters2006policy}\citep{PETERS20081180}\citep{schulman2015trust}, economics \citep{lee2001stock}\citep{moody2001learning}\citep{deng2017deep}, computer games \citep{mnih2013playing}\citep{mnih2015human}\citep{silver2016mastering} and other applications \citep{tesauro2012simulation}\citep{tesauro1995td}\citep{ernst2004power}. RL is divided into two approaches: value-based approach and policy-based approach \citep{kober2012reinforcement}. A typical value-based approach tries to obtain an optimal policy by finding optimal value functions. The value functions are updated using the immediate reward and the discounted value of the next state. Some methods based on this approach are Q-learning, SARSA, and TD-learning \citep{Sutton:1998:IRL:551283}. In contrast, the policy-based approach directly learns a parameterized policy that maximizes the cumulative discounted reward. Some techniques used for searching optimal parameters of the policy are policy gradient \citep{sutton2000policy}, expectation maximization \citep{dayan1997using},  and evolutionary algorithm \citep{moriarty1999evolutionary}. In addition, a hybrid approach of value-based approach and policy-based approach is called actor-critic \citep{konda2000actor}. Recently, RL algorithms integrated with a deep neural network (DNN) \citep{mnih2015human} achieved good performance, even better than human performance at some tasks such as Atari games \citep{mnih2015human} and Go game \citep{silver2016mastering}. However, obtaining good performance on a task consisting of multiple subtasks, such as Mazebase \citep{sukhbaatar2015mazebase} and Montezuma's Revenge \citep{mnih2013playing}, is still a major challenge for flat RL algorithms. Hierarchical reinforcement learning (HRL) \citep{sutton1999between} was developed to tackle these domains.

HRL decomposes a RL problem into a hierarchy of subproblems (subtasks) and each of which can itself be a RL problem. Identical subtasks can be gathered into one group and are controlled by the same policy. As a result, hierarchical decomposition represents the problem in a compact form and reduces the computational complexity. Various approaches to decompose the RL problem have been proposed, such as options \citep{sutton1999between}, HAMs \citep{parr1998hierarchical}\citep{parr1998reinforcement}, MAXQ \citep{dietterich2000hierarchical}, Bayesian HRL \cite{VienNLC14,VienT15,VienLC16} and some other advanced techniques \citep{barto2003recent}. All approaches commonly consider semi-Markov decision process \citep{sutton1999between} as a base theory. Recently, many studies have combined HRL with deep neural networks in different ways to improve performance on hierarchical tasks \citep{bellemare2016unifying}\citep{bacon2017option}\citep{fox2017multi}\citep{lee2017micro}\citep{vezhnevets2017feudal}\citep{durugkar2016deep}\citep{arulkumaran2016classifying}. Bacon et al. \citep{bacon2017option} proposed an option-critic architecture, which has a fixed number of intra-options, each of which is followed by a ``deep'' policy. At each time step, only one option is activated and is selected by another policy that is called ``policy over options''. DeepMind lab \citep{vezhnevets2017feudal} also proposed a deep hierarchical framework inspired by a classical framework called feudal
reinforcement learning \citep{dayan1993feudal}. Similarly, Kulkarni et al. \citep{kulkarni2016hierarchical} proposed a deep hierarchical framework of two levels in which the high-level controller produces a subgoal and the low-level controller performs primitive actions to obtain the subgoal. This framework is useful to solve a problem with multiple subgoals such as Montezuma's Revenge \citep{mnih2013playing} and games in Mazebase \citep{sukhbaatar2015mazebase}. Other studies have tried to tackle more challenging problems in HRL such as discovering subgoals \citep{chiu2011subgoal} and adaptively finding a number of options \citep{stolle2004automated}.

Though many studies have made great efforts in this topic, most of them rely on an assumption of full observability, where a learning agent can observe environment states fully. In other words, the environment is represented as a Markov decision process (MDP). This assumption does not reflect the nature of real-world applications in which the agent only observes a partial information of the environment states. Therefore, the environment, in this case, is represented as a POMDP. In order for the agent to learn in such a POMDP environment, more advanced techniques are required in order to have a good prediction over environment responses, such as maintaining a belief distribution over unobservable states; alternatively using a recurrent neural network (RNN) \citep{murphy2000survey} to summarize an observable history. Recently, there have been some studies using deep RNNs to deal with learning in POMDP environment \citep{hausknecht2015deep}\citep{egorov2015deep}.

In this paper, we develop a deep HRL algorithm for POMDP problems inspired by the deep HRL framework \citep{kulkarni2016hierarchical}. The agent in this framework makes decisions through a hierarchical policy of two levels. The top policy determines the subgoal to be achieved, while the lower-level policy performs primitive actions to achieve the selected subgoal. To learn in POMDP, we represent all policies using RNNs. The RNN in lower-level policies constructs an internal state to remember the whole states observed during the course of interaction with the environment. The top policy is a RNN to encode a sequence of observations during the execution of a selected subgoal. We highlight our contributions as follows. First, we exploit the advantages of RNNs to integrate with hierarchical RL in order to handle learning on challenging and complex tasks in POMDP. Second, this integration leads to a new hierarchical POMDP learning framework that is more scalable and data-efficient.

The rest of the paper is organized as follows. Section \ref{sec:bg} reviews the underlying knowledge such as semi-Markov decision process, partially observable Markov decision process and deep reinforcement learning. Our contributions are described in Section \ref{sec:algo}, which consists of two parts. The deep model part describes all elements in our algorithm and the learning algorithm part summarizes the algorithm in a generalized form. The usefulness of the proposed algorithms is demonstrated through POMDP benchmarks in Section \ref{sec:exp}. Finally, the conclusion is established in Section \ref{sec:con}.

\section{Background}
\label{sec:bg}

In this section, we briefly review all underlying theories that the content of this paper relies on: Markov decision process, semi-Markov decision process for hierarchical RL, partially observable Markov decision process for RL, and deep reinforcement learning.

\subsection{Markov Decision Process}

A discrete MDP models a sequence of decisions of a learning agent interacting with an environment at some discrete time scale, $t = 1, 2, \dots$. Formally, an MDP consists of a tuple of five elements $\{{\cal S},{\cal A},{\cal P},r, \gamma\}$, where $\mathcal{S}$ is a discrete state space, $\mathcal{A}$ is a discrete action space , $\mathcal{P}({s_{t + 1}},{s_t},{a_t})=p({s_{t + 1}}|{s_t},{a_t})$ is a transition function that measures the probability of obtaining a next state ${s_{t + 1}}$ given a current state-action pair $\left( {{s_t},{a_t}} \right)$, $r\left( {{s_t},{a_t}} \right)$ defines an immediate reward achieved at each state-action pair, and $\gamma \in \left( {0,1} \right)$ denotes a discount factor. MDP relies on the Markov property that a next state only depends on the current state-action pair: \[p({s_{t + 1}}|\{{s_1},{a_1},{s_2},{a_2},...,{s_t},{a_t}\}) = p({s_{t + 1}}|{s_t},{a_t}).\]
A policy of a RL algorithm is denoted by $\pi$ which is the probability of taking an action $a$ given a state $s$: $\pi = {\cal P}(a|s)$, the goal of a RL algorithm is to find an optimal policy $\pi^*$ in order to maximize the expected total discounted reward as follows
\begin{align}
J(\pi ) = \mathbb{E}\bigg[\sum\limits_{t = 0}^\infty {{\gamma ^t}r({s_t},{a_t})}\bigg].
\end{align}

\subsection{Semi-Markov Decision Process for Hierarchical Reinforcement Learning}

Learning over different levels of policy is the main challenge for hierarchical tasks. The semi-Markov decision process (SMDP) \citep{sutton1999between}, which is as an extension of MDP,  was developed to deal with this challenge. In this theory, the concept ``options'' is introduced as a type of temporal abstraction. An ``option'' $\xi \in \Xi$ is defined by three elements: an option's policy $\pi$, a termination condition $\beta$, and an initiation set ${\cal I}\subseteq {\cal S}$ denoted as the set of states in the option. In addition, a policy over options $\mu(\xi|s)$ is introduced to select options. An option is executed as follows. First, when an option's policy $\pi ^ \xi$ is taken, the action $a _t$ is selected based on $\pi ^ \xi$. The environment then transitions to state $s _{t+1}$. The option either terminates or continues according to a termination probability $\beta^\xi(s_{t+1})$. When the option terminates, the agent can select a next option based on $\mu(\xi|s_{t+1})$. The total discounted reward received by executing option $\xi$ is defined as 
\begin{align}
{\cal R} _s ^\xi = \mathbb{E}\bigg[\sum\limits_{i = t}^{t+k-1} {\gamma ^{i-t}}r^\xi ({s_i},{a_i})\bigg].
\end{align}
The multi-time state transition model of option $\xi$ \cite{white1976procedures}, which is initiated in state $s$ and terminate at state $s'$ after $k$ steps, has the formula
\begin{align}
{\cal P} _{ss'} ^\xi = \sum\limits_{k = 1}^\infty {\cal P}^\xi (s', k|s,\xi) \gamma ^k.
\end{align}
where ${\cal P}^\xi (s', k|s,\xi)$ is the joint probability of ending up at $s'$ after $k$ steps if taking option $\xi$ at state $s$. Given this, we can write the Bellman equation for the value function of a policy $\mu$ over options:
\begin{align}
{\cal V} ^ \mu (s) = \sum\limits_{\xi \in \Xi} \mu(\xi|s) \bigg[  {\cal R} _s ^\xi + \sum\limits_{s'} {\cal P} _{ss'}^\xi {\cal V} ^ \mu (s')\bigg]
\end{align}
and the option-value function:
\begin{align}
{\cal Q} ^\mu (s, \xi) = {\cal R} _s ^\xi + \sum\limits_{s'} {\cal P} _{ss'}^\xi \sum\limits_{\xi' \in {\Xi}} \mu (\xi'|s') {\cal Q} ^ \mu (s', \xi').
\end{align}
Similarly, the corresponding optimal Bellman equations are as follows:
\begin{align}
{\cal V} ^* (s) = \max\limits_{\xi \in {\Xi}} \bigg[  {\cal R} _s ^\xi + \sum\limits_{s'} {\cal P} _{ss'}^\xi {\cal V} ^* (s')\bigg]
\end{align}
\begin{align}
{\cal Q} ^* (s, \xi) = {\cal R} _s ^\xi + \sum\limits_{s'} {\cal P} _{ss'}^\xi  \max\limits_{\xi' \in {\Xi}} {\cal Q} ^* (s', \xi').
\end{align}
The optimal Bellman equation can be computed using synchronous value iteration (SVI) \citep{sutton1999between}, which iterates the following steps for every state:
\begin{align}
{\cal V} (s) = \max\limits_{\xi \in {\Xi}} \bigg[  {\cal R} _s ^\xi + \sum\limits_{s'} {\cal P} _{ss'}^\xi {\cal V} (s', \xi')\bigg].
\end{align}

When the option model is unknown, ${\cal Q} _t (s, \xi)$ can be estimated using a Q-learning algorithm with an estimation formula: 
\begin{align}
{\cal Q} (s, \xi) \leftarrow  {\cal Q} (s, \xi) + \alpha \bigg[ r(s,\xi(s)) + \gamma ^k \max\limits_{\xi' \in {\Xi}} {\cal Q} (s', \xi') - {\cal Q} (s, \xi) \bigg],
\end{align}
where $\alpha$ denotes the learning rate, $k$ denotes the number of time steps elapsing between $s$ and $s'$ and $r$ denotes an intermediate reward if $\xi(s)$ is a primitive action, otherwise $r$ is the total reward when executing option $\xi(s)$.

\subsection{Partially observable Markov Decision Process in Reinforcement Learning}

In many real-world tasks, the agent might not have full observability over the environment. In principle, those tasks can in principle be formulated as a POMDP which is defined as a tuple of six components $\{{\cal S}, {\cal A}, {\cal P}, r, {\Omega}, {\cal Z}\}$, where ${\cal S}, {\cal A}, {\cal P}, r$ are the state space, action space, transition function and reward function, respectively as in a Markov decision process.  ${\Omega}$ and ${\cal Z}$ are observation space and observation model, respectively. If the POMDP model is known, the optimal approach is to maintain a hidden state $b_t$ called belief state. The belief defines the probability of being in state $s$ according to its history of actions and observations. Given a new action and observation, belief updates are performed using the Bayes rule \citep{kaelbling1998planning} and defined as follows: 
\begin{align}
b'(s') \propto {\cal Z}(s',a, o) \sum\limits_{s \in {\cal S}} {\cal P}(s,a,s')b(s).
\end{align}
However, exact belief updates require a high computational cost and expensive memory \citep{murphy2000survey}. Another approach is using a Q-learning agent with function approximation, which uses Q-learning algorithm for updating the policy. Because updating the Q-value from an observation can be less accurate than estimating from a complete state, a better way would be that a POMDP agent using Q-Learning uses the last $k$ observations as input to the policy. Nevertheless, the problem with using a finite number of observations is that key-event observations far in the past would be neglected in future decisions. For this reason, a RNN is used to maintain a long-term state, as in \citep{hausknecht2015deep}. Our model using RNNs at different levels of the hierarchical policies is expected to take advantage in POMDP environments.

\subsection{Deep Reinforcement Learning}

Recent advances in deep learning \citep{Goodfellow-et-al-2016} are widely applied to reinforcement learning to form deep reinforcement learning. A few years ago, reinforcement learning still used ``shallow'' policies such as tabular, linear, radial basis network, or neural networks of few layers. The shallow policies contain many limitations, especially in representing highly complex behaviors and the computational cost of updating parameters. In contrast, deep neural networks in deep reinforcement learning can extract more information from the raw inputs by pushing them through multiple neural layers such as multilayer perceptron layer (MLP) and convolutional layer (CONV). Multiple layers in DNNs can have a lot of parameters allowing them to represent highly non-linear problems. Deep Q-Network (DQN) has been proposed recently by Google Deepmind \citep{mnih2015human}, that opens a new era of deep reinforcement learning and influenced most later studies in deep reinforcement learning. In term of architecture, a Q-network parameterized by $\theta$, e.g., $Q(s, a | \theta)$ is built on a convolutional neural network (CNN) which receives an input of four images of size $84 \times 84$ and is processed by three hidden CONV layers. The final hidden layer is a MLP layer with 512 rectifier units. The output layer is a MLP layer, which has the number of outputs equal to the number of actions. In term of the learning algorithm, DQN learns Q-value functions iteratively by updating the Q-value estimation via the temporal difference error:
\begin{align}
Q(s,a) := Q(s,a) + \alpha (r + \gamma \max \limits _{a'} Q(s',a') - Q(s,a)).
\end{align}
In addition, the stability of DQN also relies on two mechanisms. The first mechanism is \textit{experience replay memory}, which stores transition data in the form of tuples $\{s, a, s', r\}$. It allows the agent to uniformly sample from and train on previous data (off-policy) in batch, thus reducing the variance of learning updates and breaking the temporal correlation among data samples. The second mechanism is the target Q-network, which is parameterized by $\theta'$, e.g., $\hat{Q}(s,a|\theta')$, and is a copy of the main Q-network. The target Q-network is used to estimate the loss function as follows:
\begin{align}
{\cal L} = \mathbb{E} \bigg[ (y - Q(s,a |\theta))^2\bigg],
\end{align}
where $y = r + \gamma \max\limits_{a'} \hat{Q} (s',a' | \theta')$. Initially, the parameters of the target Q-network are the same as in the main Q-network. However, during the learning iteration, they are only updated at a specified time step. This update rule causes the target Q-network to decouple from the main Q-network and improves the stability of the learning process.

Many other models based on DQNs have been developed such as Double DQNs \citep{van2016deep}, Dueling DQN \citep{wang2015dueling}, and Priority Experiment Replay \citep{schaul2015prioritized}. On the other hand, deep neural networks can be integrated into other methods rather than estimating Q-values, such as representing the policy in policy search algorithms \citep{schulman2015trust}, estimating advantage function \citep{gu2016continuous}, or mixed actor network and critic network \citep{lillicrap2015continuous}. 

Recently, researchers have used RNNs in reinforcement learning to deal with POMDP domains. RNNs have been successfully applied to domains, such as natural language processing and speech recognition, and are expected to be advantageous in the POMDP domain, which needs to process a sequence of observation rather than a single input. Our proposed method uses RNNs not only for solving the POMDP domain but also for solving these domains in a hierarchical form of reinforcement learning.

\section{Hierarchical Deep Recurrent Q-Learning in Partially Observable Semi-Markov Decision Process}
\label{sec:algo}

In this section, we describe hierarchical recurrent Q learning algorithm (hDRQN), our proposed framework. First, we describe the model of hDRQN and explain the way to learn this model. Second, we describe some components of our algorithm such as sampling strategies, subgoal definition, extrinsic and intrinsic reward functions. We rely on partially observable semi-Markov decision process (POSMDP) settings, where the agent follows a hierarchical policy to solve POMDP domains. The setting of POSMDP \cite{white1976procedures,VienT15} is described as follows. The domain is decomposed into multiple subdomains. Each subdomain is equivalent to an option $\xi$ in the SMDP framework and has a subgoal $g \in \Omega$, that needs to be achieved before switching to another option. Within an option, the agent observes a partial state $o \in \Omega$, takes an action $a \in {\cal A}$, receives an extrinsic reward $r^{ex} \in \mathbb{R}$, and transits to another state $s'$ (but the agent only observes a part of the state $o' \in \Omega$). The agent executes option $\xi$ until it is terminated (either the subgoal $g$ is obtained or the termination signal is raised). Afterward, the agent selects and executes another option. The sequence of options is repeated until the agent reaches the final goal. In addition, for obtaining subgoal $g$ in each option, an intrinsic reward function $r^{in} \in \mathbb{R}$ is maintained. While the objective of a subdomain is to maximize the cumulative discounted intrinsic reward, the objective of the whole domain is to maximize the cumulative discounted extrinsic reward.

Specifically, the belief update given a taken option $\xi$, observation $o$ and current belief $b$ is defined as
  \begin{align*}
    b'(s') \propto \sum_{k=1}^{\infty} \gamma^{k-1} \sum_{s} {\cal P}(s',o,k|s,\xi) b(s)
  \end{align*}
  where ${\cal P}(s',o,k|s,\xi)$ is a joint transition and observation function of the underlying POSMDP model on the environment.

  We adopt a similar notation from the two frameworks MAX-Q \cite{dietterich2000hierarchical} and Options \cite{sutton1999between}, to describe our problem. We denote $Q^M(o_t,\xi|\theta^M)$ as the Q-value function of the meta-controller at state $o_t,\theta^M$ (in which we use a RNN to encode past observation that has been encoded in its weights $\theta^M$) and option (macro-action) $\xi$ (assuming $\xi$ has a corresponding subgoal $g_t$). We note that  the pair $\{o_t,\theta^M\}$ represents the belief or history at time $t$, $b_t$ (we will use them inter-changeably: $Q^M(b,\xi)$ or $Q^{M}(o,\xi|\theta^M)$).
  
The multi-time observation model of option $\xi$ \cite{white1976procedures}, which is initiated in belief $b$ and observe $o$, has a formula as follows
\begin{align}
{\cal P} (o|b,\xi) = \sum\limits_{k = 1}^\infty \sum_{s'} \sum_s  \gamma ^k {\cal P} (s',o, k|s,\xi)b(s)
\end{align}
Given above, we can write the Bellman equation for the value function of the meta-controller policy $\pi$  over options as follows
\begin{align}
{\cal V} ^ M (b) = \sum_{\xi}\pi(\xi|b) \left[  {\cal R} _b ^\xi + \sum\limits_{o'} {\cal P}(o'|b,\xi)  {\cal V} ^ M (b') \right]
\end{align}
and the option-value function,
\begin{align}
{\cal Q} ^ M (b,\xi) = {\cal R} _b ^\xi + \sum\limits_{o'} {\cal P}(o'|b,\xi)\sum\limits_{\xi' \in {\Xi}} \mu (\xi'|b') {\cal Q} ^ M (b', \xi') 
\end{align}

Similarly to the MAX-Q framework, the reward term $R_b^{\xi}$ is the total reward collected by executing the option $\xi$ and defined as $V^{\xi}(b)$. Its corresponding Q-value function is defined as $Q^{S}(b,a)$ (the value function for sub-controllers). In the use of RNN, we also denote $Q^{S}(b,a)$ as $Q^{S}(\{o_t,g_t\},a|\theta^S)$ in which $\theta^S$ is the weights of the sub-controller network that encodes previous observations, and $\{o_t,g_t\}$ denote the observation input to the sub-controller.

\begin{figure}[htbp]
    \centering
    \begin{tabular}{cc}
    \begin{tabular}{@{}c@{}}
        \subfloat[Framework 1\label{fig:framework_1}]{
            \includegraphics[width=.4\linewidth]{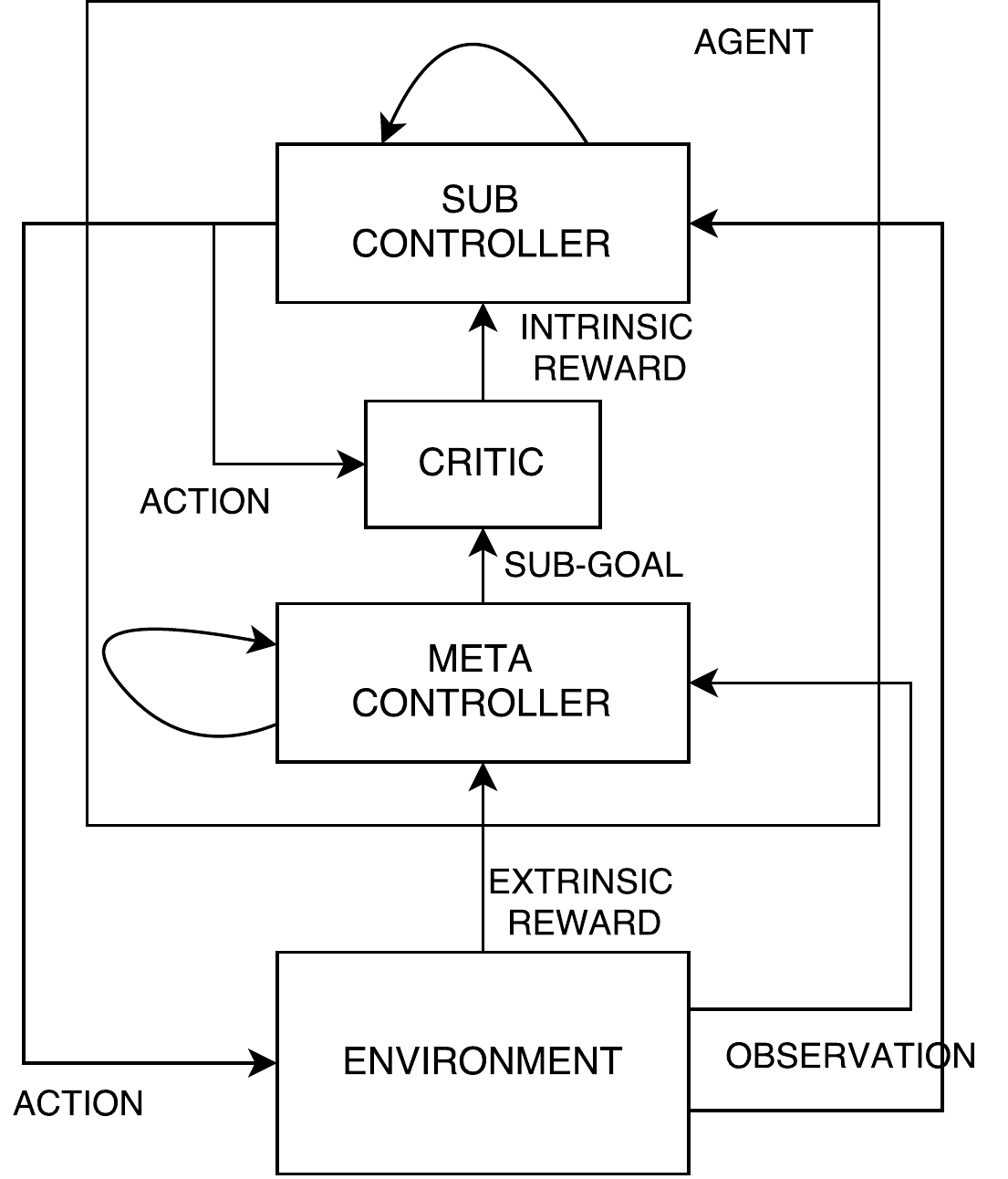}}
        \end{tabular}
    &
    \begin{tabular}{@{}c@{}}
        \subfloat[Framework 2\label{fig:framework_2}]{
            \includegraphics[width=.4\linewidth]{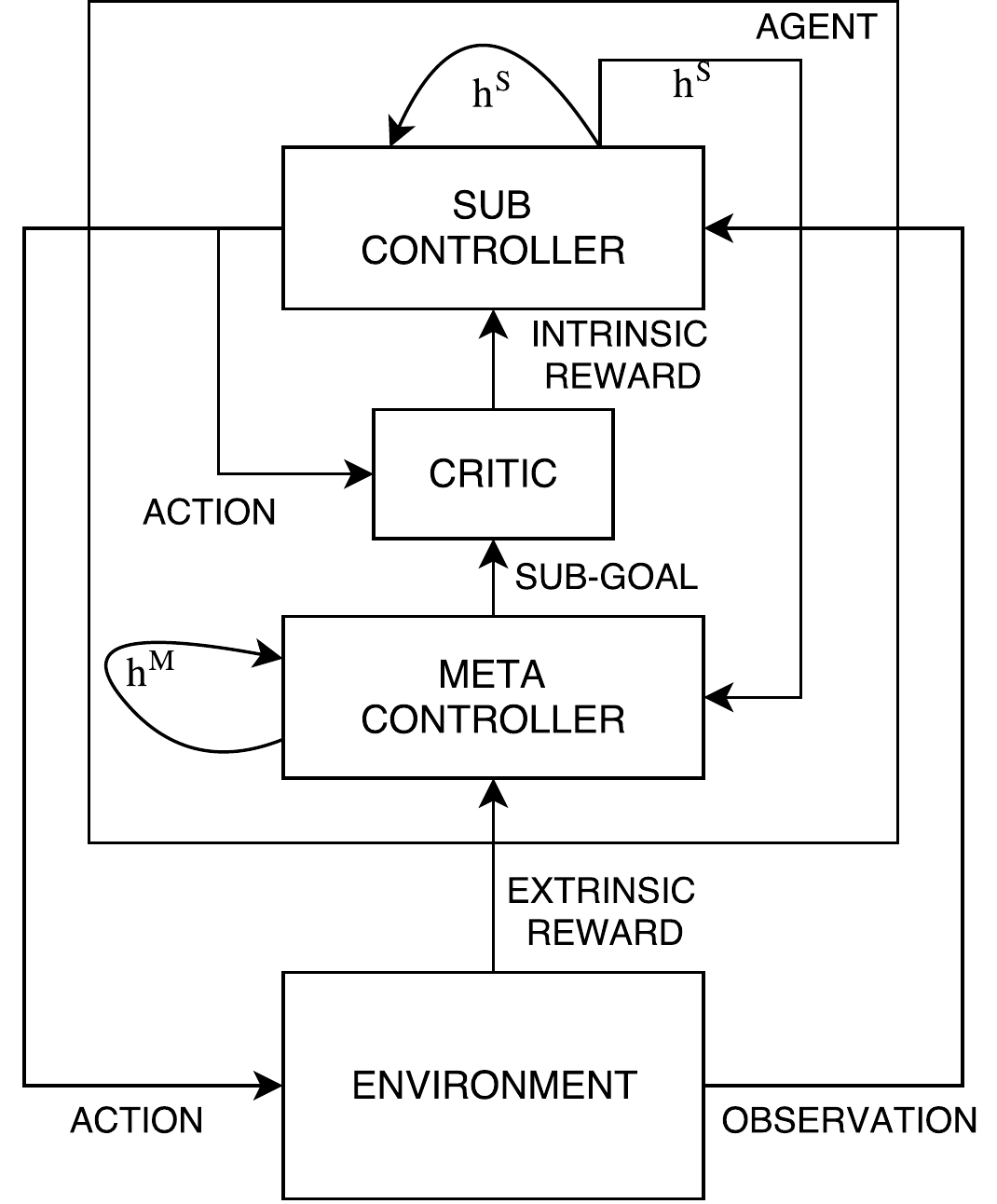}}
        \end{tabular}
    \end{tabular}
    \caption{Hierarchical deep recurrent Q-network frameworks}\label{fig:framework}
\end{figure}

Our hierarchical frameworks are illustrated in Fig. \ref{fig:framework}. The framework in Fig. \ref{fig:framework_1} is inspired by a related idea in \citep{kulkarni2016hierarchical}. The difference is that our framework is built on two deep recurrent neural policies: a meta-controller and a sub-controller. The meta-controller is equivalent to a ``policy over subgoals'' that receives the current observation $o_t$ and determine new subgoal $g_t$. The sub-controller is equivalent to the option's policy, which directly interacts with the environment by performing action $a_t$. The sub-controller receives both $g_t$ and $o_t$ as inputs to the deep neural network. Each controller contains an arrow (Fig. \ref{fig:framework}) pointed to itself that indicates that the controller employs recurrent neural networks. In addition, an internal component called ``critic'' is used to determine whether the agent has achieved subgoal or not and to produce an intrinsic reward that is used to learn the sub-controller. In contrast to the framework in Fig. \ref{fig:framework_1}, the framework in Fig. \ref{fig:framework_2} does not use the current observation to determine the subgoal in the meta-controller but instead uses the last hidden state $h^S_{t}$ of the sub-controller. The last hidden state that is inferred from a sequence of observations of the sub-controller is expected to contribute to the meta-controller to correctly determine the next subgoal.

\begin{figure*}[htbp]
    \centering
    \adjustbox{valign=b}{
    \begin{tabular}{@{}c@{}}
       \subfloat[Framework 1
           \label{fig:framework_detail_1}]{
            \includegraphics[width=0.8\linewidth]{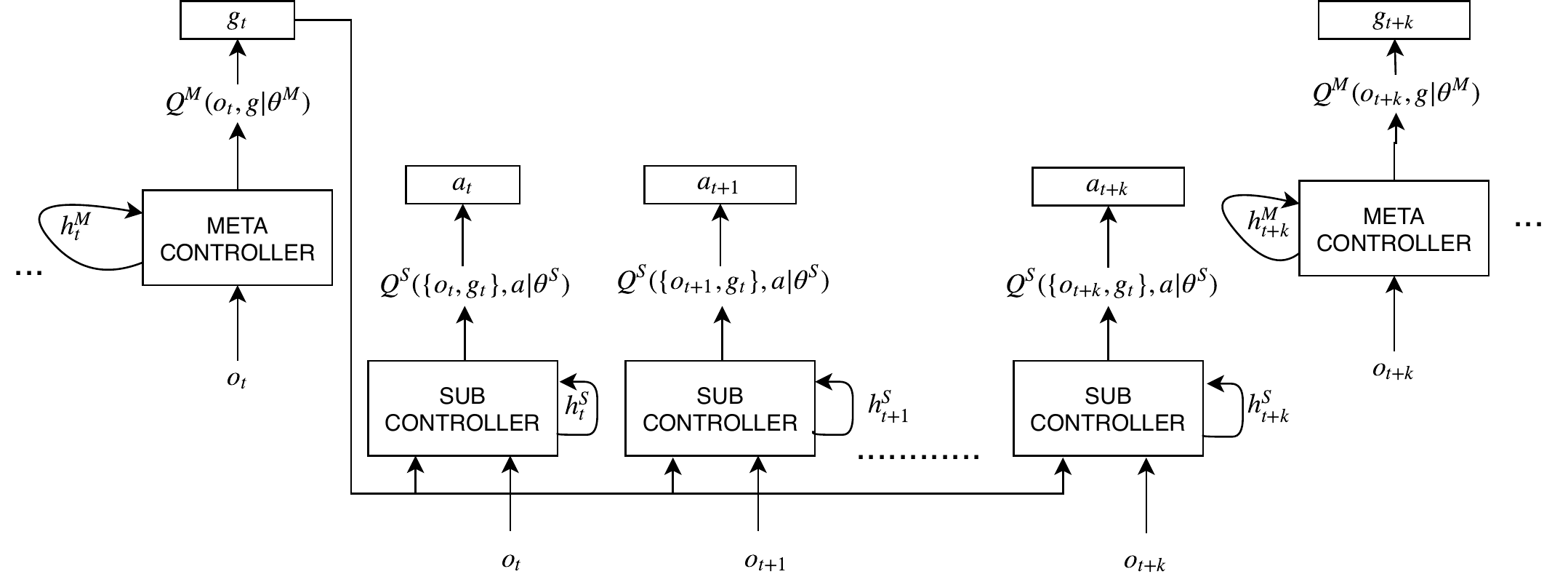}} \\
       \subfloat[Framework 2
            \label{fig:framework_detail_2}]{
             \includegraphics[width=0.8\linewidth]{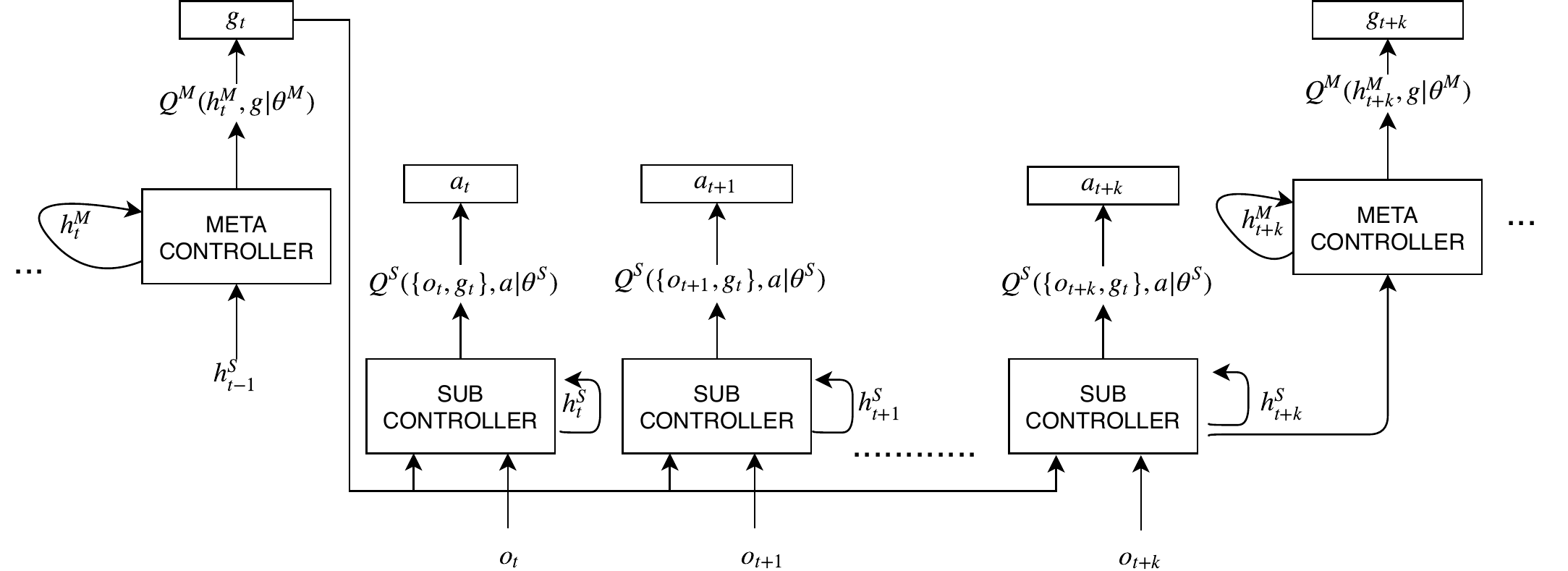}}
        \end{tabular}}
    \caption{Procedure of Hierarchical Deep Recurrent Q Learning (hDRQN). The framework is based on \citep{kulkarni2016hierarchical}}
    \label{fig:framework_detail}
\end{figure*}

As mentioned in the previous section, RNNs is used in our framework in order to enable learning in POMDP. Particularly, CNNs is used to learn a low-dimensional representation for image inputs, while RNN is used in order to capture a temporal relation of observation data. Without using a recurrent layer as in RNNs, CNNs cannot accurately approximate the state feature from observations in POMDP domain. The procedure of the agent is clearly illustrated in Fig. \ref{fig:framework_detail}. The meta-controller and sub-controller use the Deep Recurrent Q Network (DRQN), as described in \citep{hausknecht2015deep}. Particularly, at step $t$, the meta-controller takes an observation $o_t$ from the environment (framework 1) or the last hidden state of sub-controller generated by the previous sequence of observations (framework 2), extracts state features through some deep neural layers, internally constructs a hidden state $h^M_t$, and produces the $Q$ subgoal values $Q^M(o_t, g_t | \theta^M)$. The $Q$ subgoal values then are used to determine the next subgoal $g_{t+k}$. Similarly, the sub-controller receives both observation $o_t$ and subgoal $g_t$, extracts their feature, constructs a hidden state $h^S_t$,  and produces $Q$ action values $Q^S(\{o_t, g_t\}, a_t  | \theta^S)$, which are used to determine the next action $a_{t+1}$. Those explanations are formalized into the following equations:
\begin{align}
\Phi^M = \begin{cases}
f^{extract}(o_t) & \text{framework 1}\\
f^{extract}(h^S) & \text{framework 2}
\end{cases} \\
h^M_t, Q^M(o_t, g_t | \theta^M) = f^M(\Phi^M, h^M_{t-1}) \\
\Phi^S = f^{extract}(o_t, g_t) \\
h^S_t, Q^S(\{o_t, g_t\}, a_t | \theta^S) = f^S(\Phi^S, h^S_{t-1}),
\label{eq:formula}
\end{align}
where $\Phi^M$ and $\Phi^S$ are the features after the extraction process of the meta-controller and sub-controller, respectively. $f^M$ and $f^S$ are the respective recurrent networks of the meta-controller and sub-controller. They receive the state features and a hidden state, and then provide the next hidden state and $Q$ value function.

\begin{figure}[htbp]
    \centering
    \adjustbox{valign=b}{
    \begin{tabular}{@{}c@{}}
       \subfloat[The network for framework 1
           \label{fig:network_model_1}]{
            \includegraphics[width=0.9\linewidth]{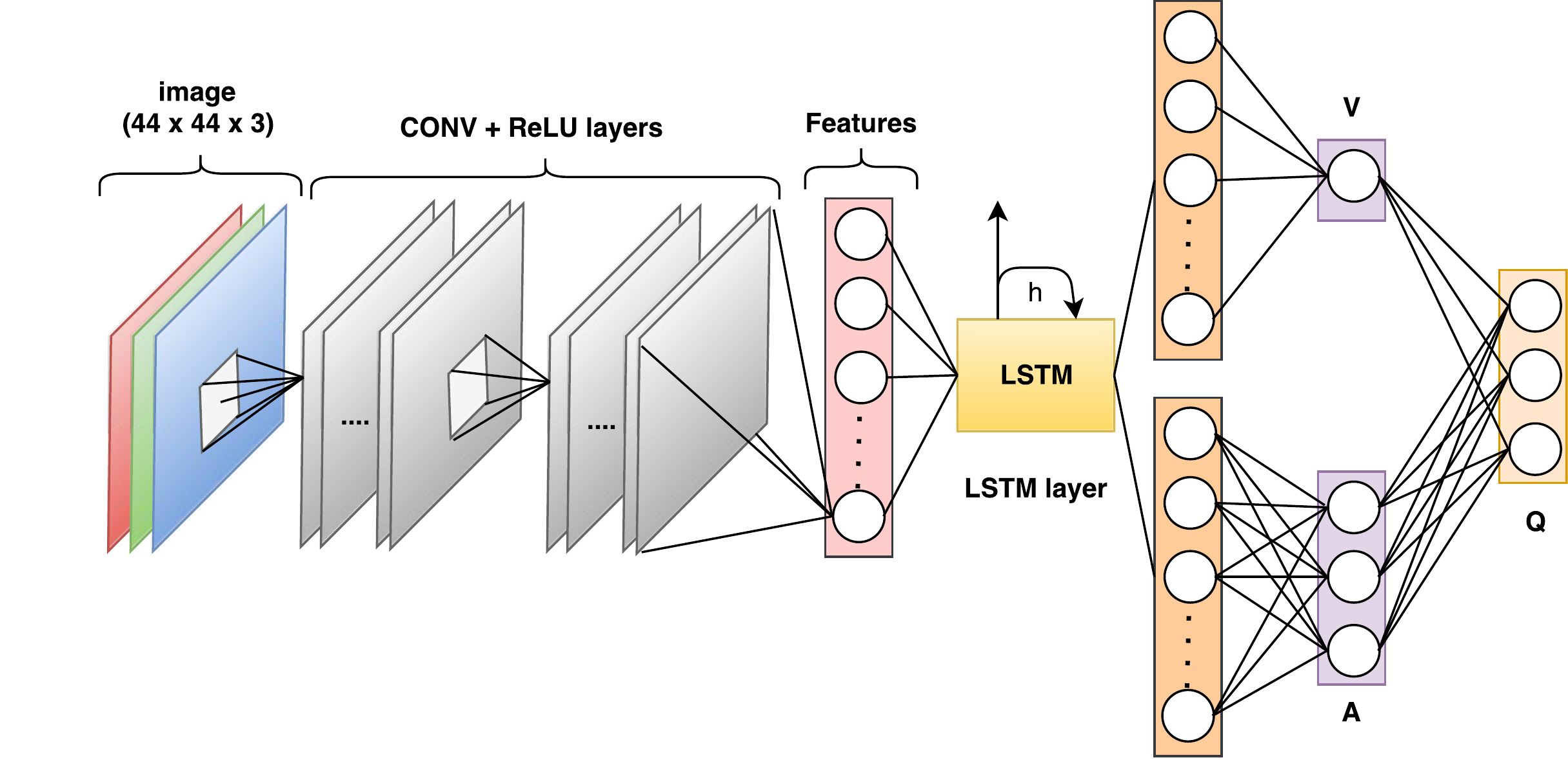}} \\
       \subfloat[Meta-controller network in framework 2
            \label{fig:network_model_2}]{
             \includegraphics[width=0.9\linewidth]{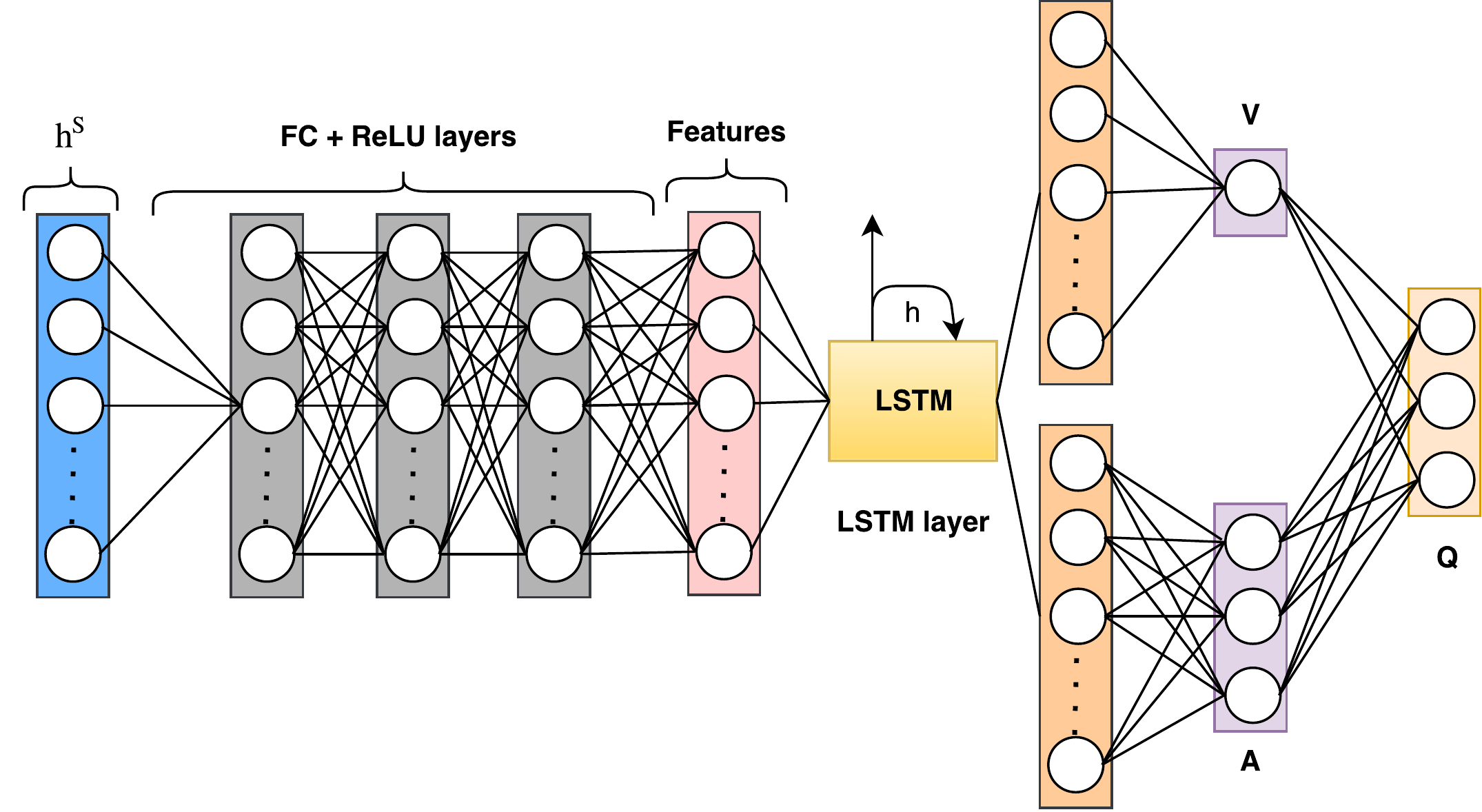}}
        \end{tabular}}
    \caption{Network models}
    \label{fig:network_model}
\end{figure}

The networks for controllers are illustrated in Fig. \ref{fig:network_model}. For framework 1, we use the networks demonstrated in Fig. \ref{fig:network_model_1} for both the meta-controller and sub-controller. A sequence of four CONV layers and ReLU layers interleaved together is used to extract information from raw observations. A RNN layer, especially LSTM, is employed in front of the feature to memorize information from previous observations. The output of the RNN layer is split into Advantage stream $A$ and Value stream $V$ before being unified to the Q-value. This architecture inherits from Dueling architecture \citep{wang2015dueling}, effectively learning states without having to learn the effect of an action on that state. For framework 2, we use the network in Fig. \ref{fig:network_model_1} for the sub-controller and use the network in Fig. \ref{fig:network_model_2} to the meta-controller. The meta-controller in framework 2 uses the state that is the internal hidden state from the sub-controller. By passing through four fully connected layers and ReLU layers, features of the meta-controller are extracted. The rest of the network is the same as the network for framework 1.

\subsection{Learning Model}
\label{sec:learn}
We use a state-of-the-art Double DQN to learn the parameters of the network.  Particularly, the controllers estimate the following $Q$ value function: 
\begin{multline}
Q^M(\hat{o}_t, g_t) = Q^M(\hat{o}_t, g_t) + \\ \alpha (r^{ex} + \gamma^k \max \limits _{g_{t+k}} Q^M(\hat{o}_{t+k},g_{t+k}) - Q^M(\hat{o}_t,g_t))
\end{multline} and 
\begin{multline}
Q^S(\{o_t, g_t\}, a_t) =  Q^S(\{o_t, g_t\}, a_t) + \\ \alpha (r^{in} + \gamma \max \limits _{a_{t+1}} Q^S(\{o_{t+1}, g_t\}, a_{t+1}) - Q^S(\{o_t, g_t\}, a_t))
\end{multline}
where $\hat{o}$ can be a direct observation or an internal hidden state generated by the sub-controller. Let $\theta^M$ and $\theta^S$ be parameters (weights and bias) of the deep neural networks, that parameterize the networks $Q^M(\hat{o}_t, g_t)$ and $Q^S(\{o_t, g_t\}, a_t)$, correspondingly, e.g. $Q^M(\hat{o}_t, g_t) = Q^M(\hat{o}_t, g_t | \theta^M)$ and $Q^S(\{o_t, g_t\}, a_t) = Q^S(\{o_t, g_t\}, a_t | \theta^S)$. Then, $Q^M$ and $Q^S$ are trained by minimizing the loss function $L^M$ and $L^S$, correspondingly. $L^M$ can be formulated as: 
\begin{align}
L^M =  \mathbb{E}_{(o,g,o',g',r^{ex}) \sim \cal {M}^M} \big[ y^M_i - Q^M(o,g | \theta^M)\big],
\end{align}
where $\mathbb{E}$ denotes the expectation over a batch of data which is uniformly sampled from an experience replay $M^M$, $i$ is the iteration number in the batch of samples and 
\begin{align}
y^M_i = r^{ex} + \gamma Q^{M'}(o', \operatorname*{argmax}_{g'} Q^{M}(o',g' | \theta^{M}) |\theta^{M'}).
\end{align}
Similarly, the formula of $L^S$ is 
\begin{align}
L^S =  \mathbb{E}_{(o,g,a,r^{in}) \sim \cal {M}^S} \big[ y^S_i - Q^S(\{o,g\},a | \theta^S)\big],
\end{align}
where 
\begin{align}
y^S_i = r^{in} + \gamma Q^{S'}(\{o', g\}, \operatorname*{argmax}_{a'} Q^S (\{o',g\},a' | \theta^{S}) | \theta^{S'}).
\end{align}
$\cal {M}^S$ is experience replay that stores the transition data from sub-controller, and $Q^{S'}$ is the target network of $Q^S$. Intuitively, in contrast to DQN which uses the maximum operator for both selecting and evaluating an action, Double DQN separately uses the main Q network to greedily estimate the next action and the target Q network to estimate the value function. This method has been shown to achieve better performance than DQN on Atari games \citep{van2016deep}. 

\subsection{Minibatch sampling strategy}
\label{sec:sample}

For updating RNNs in our model, we need to pass a sequence of samples. Particularly, some episodes from the experience replay are uniformly sampled and are processed from the beginning of the episode to the end of the episode. This strategy called Bootstrapped Sequential Updates \citep{hausknecht2015deep}, is an ideal method to update RNNs because their hidden state can carry all information through the whole episode. However, this strategy is computationally expensive in a long episode, which can contain many time steps. Another approach proposed in \citep{hausknecht2015deep} has been evaluated to achieve the same performance as Bootstrapped Sequential Updates, but reduce the computational complexity. The strategy is called Bootstrapped Random Updates and has a procedure as follows. This strategy also randomly selects a batch of episodes from the experience replay. Then, for each episode, we begin at a random transition and select a sequence of $n$ transitions. The value of $n$ that affects to the performance of our algorithm is analyzed in section \ref{sec:exp}. We apply the same procedure of Bootstrapped Random Updates to our algorithm.

In addition, the mechanism explained in \citep{lample2017playing} is also applied. That study explains a problem when updating DRQN: using the first observations in a sequence of transitions to update the $Q$ value function might be inaccurate. Thus, the solution is to use the last observations to update DRQN. Particularly, our method uses the last $\frac{n}{2}$ transitions to update the Q-value.

\subsection{Subgoal Definition}

Our model is based on the ``option'' framework. Learning an option is using flat deep RL algorithms to achieve a subgoal of that option. However, discovering subgoals among existing states in the environment is still one of the challenges in hierarchical reinforcement learning. For simplifying the model, we assume that a set of pre-defined subgoals is provided in advance. The pre-defined subgoals based on object-oriented MDPs \citep{diuk2008object}, where entities or objects in the environment are decoded as subgoals. 

\subsection{Intrinsic and Extrinsic Reward Functions}
\label{sec:extrinsic}

\begin{figure}[htbp]
\centering
\includegraphics[width=0.3\textwidth]{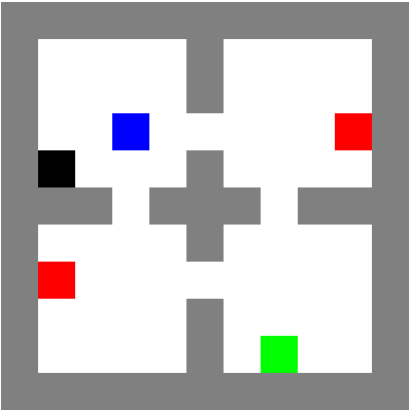}
\caption{Example domain for illustrating the notions of intrinsic and extrinsic motivation}
\label{fig:example}
\end{figure}

Traditional RL accumulates all reward and penalty in a reward function, which is hard to learn in a specified task in a complex domain. In contrast, hierarchical RL introduces the notions of intrinsic reward function and an extrinsic reward function. Initially, intrinsic motivation is based on psychology, cognitive science, and neuroscience \citep{ryan2000intrinsic} and has been applied to hierarchical RL \citep{stout2005intrinsically}\citep{barto2004intrinsically}\citep{singh2010intrinsically}\citep{frank2014curiosity}\citep{mohamed2015variational}\citep{schmidhuber2010formal}. Our framework follows the model of intrinsic motivation in \citep{barto2004intrinsically}. Particularly, within an option (or skill), the agent needs to learn an option's policy (sub-controller in our framework) to obtain a subgoal (a salient event) under reinforcing of an intrinsic reward while for the overall task, a policy over options (meta-controller) is learned to generate a sequence of subgoals while reinforcing an extrinsic reward. Defining ``good'' intrinsic reward and extrinsic reward functions is still an open question in reinforcement learning, and it is difficult to find a reward function model that is generalized to all domains. To demonstrate some notions above, Fig. \ref{fig:example} describes the domain of multiple goals in four-rooms which is used to evaluate our algorithm in Section \ref{sec:exp}. The four-rooms contain a number of objects: an agent (in black), two obstacles (in red) and two goals (in blue and green). These objects are randomly located on the map. At each time step, the agent has to follow one of the four actions: top, down, left or right, and has to move to the goal location in a proper order: the blue goal first and then the green goal. If the agent obtains all goals in the right order, it will receive a big reward; otherwise, it will only receive a small reward. In addition, the agent has to learn to avoid the obstacles if it does not want to be penalized. For this example, the salient event is equivalent to reaching the subgoal or hitting the obstacle. In addition, there are two skills the agent should learn. One is moving to the goals while correctly avoiding the obstacles, and the second is selecting the goal it should reach first. The intrinsic reward for each skill is generated based on the salient events encounters while exploring the environment. Particularly, for reaching the goal, the intrinsic reward includes the reward for reaching goal successful and the penalty if the agent encounters an obstacle. For reaching the goals in order, the intrinsic reward includes a big reward if the agent reaches the goals in a proper order and a small reward if the agent reaches the goal in an improper order. A detailed explanation of intrinsic and extrinsic rewards for this domain is included in Section \ref{sec:exp}.

\subsection{Learning Algorithm}

\begin{algorithm}[htbp]
\begin{algorithmic}[1]
\REQUIRE {
\STATE POMDP $M = \{{\cal S}, {\cal A}, {\cal P}, r, {\Omega}, {\cal Z}\}$
\STATE Meta-controller with the network $Q^M$ (main) and $Q^{M'}$ (target) parameterized by $\theta^M$ and $\theta^{M'}$, respectively 
\STATE Sub-controller with the network $Q^S$ (main) and $Q^{S'}$ (target) parameterized by $\theta^S$ and $\theta^{S'}$, respectively
\STATE Exploration rate $\epsilon^M$ for meta-controller and $\epsilon^S$ for sub-controller
\STATE Experience replay memories $M^M$ and $M^S$ of meta-controller and sub-controller respectively
\STATE A pre-defined set of subgoals $\cal G$
\STATE $f^M$ and $f^S$ are recurrent networks of meta-controller and sub-controller respectively
}
\ENSURE
\STATE \textbf{Initialize:}
\begin{itemize}
  \item Experiences replay memories $M^M$ and $M^S$\;
  \item Randomly initialize  $\theta^M$ and $\theta^S$\;
  \item Assign value to the target networks $\theta^{M'} \leftarrow \theta^{M}$ and $\theta^{S'} \leftarrow \theta^{S}$ \;
  \item $\epsilon^M \leftarrow 1.0$ and decreasing to $0.1$ \;
  \item $\epsilon^S \leftarrow 1.0$ and decreasing to $0.1$ \;
\end{itemize}
\FOR{$k = 1,2, \ldots K$}
    \STATE \textbf{Initialize:} the environment and get the start observation $o$ \;
    \STATE \textbf{Initialize:} hidden states $h^M \leftarrow \mathbf{0}$\;
    \WHILE{$o$ is \textbf{not} terminal}
        \STATE \textbf{Initialize:} hidden states $h^S \leftarrow \mathbf{0}$\;
        \STATE \textbf{Initialize:} start observations $o_0 \leftarrow \hat{o}$ where $\hat{o}$ can be observation $o$ or hidden state $h^S$\;
        \STATE \textbf{Determine subgoal:} $g, h^M \leftarrow $ \newline $EPS\_GREEDY(\hat{o}, h^M, {\cal G}, \epsilon^M, Q^M, f^M)$ \;
        \WHILE{$o$ is \textbf{not} terminal \textbf{and} $g$ is \textbf{not} reached} 
            \STATE \textbf{Determine action:} $a, h^S \leftarrow $  \newline $EPS\_GREEDY(\{o, g\}, h^S, {\cal A}, \epsilon^S, Q^S, f^S)$ \;
            \STATE \textbf{Execute} action $a$, receive reward $r$, extrinsic reward $r^{ex}$, intrinsic reward $r^{in}$, and obtain the next state $s'$ \;
            \STATE \textbf{Store transition} $\{ \{o,g\},a,r^{in},\{o',g'\} \}$ in $M^S$ \;
            \STATE \textbf{Update sub-controller}  \newline $SUB\_UPDATE(M^S, Q^S, Q^{S'})$ \;
            \STATE \textbf{Update meta-controller}  \newline $META\_UPDATE(M^M, Q^M, Q^{M'})$ \;
            \STATE \textbf{Transition to next observation} $o \leftarrow o'$ \;
        \ENDWHILE
        \STATE \textbf{Store transition} $\{ o_0, g, r^{ex}_{total}, \hat{o}' \}$ in $M^S$ where $\hat{o}'$ can be observation $o'$ or the last hidden state $h^S$\;
    \ENDWHILE
    
    \STATE \textbf{Anneal} $\epsilon^M$ and $\epsilon^S$\;
\ENDFOR
\caption{hDRQN in POMDP}
\label{alg:hDRQN}
\end{algorithmic}
\end{algorithm}
In this section, our contributions are summarized through pseudo-code Algorithm \ref{alg:hDRQN}. The algorithm learns four neural networks: two networks for meta-controller ($Q^M$ and $Q^{M'}$) and two networks for sub-controller ($Q^S$ and $Q^{S'}$). They are parameterized by $\theta^M$, $\theta^{M'}$, $\theta^S$ and $\theta^{S'}$. The architectures of the networks are described in Section \ref{sec:algo}. In addition, the algorithm separately maintains two experience replay memories $M^M$ and $M^S$ to store transition data from meta-controller and sub-controller, respectively. Before starting the algorithm, the parameters of the main networks are randomly initialized and are copied to the target networks. $\epsilon^M$ and $\epsilon^S$ are annealed from 1.0 to 0.1, which gradually increase control of controllers. The algorithm loops through a specified number of episodes (Line 9) and each episode is executed until the agent reaches the terminal state. To start an episode, first, a starting observation $o_0$ is obtained (Line 10). Next, hidden states, which are inputs to RNNs, must be initialized with zero values (Line 11 and Line 13) and are updated during the episode (Line 15 and Line 17).  Each subgoal is determined by passing observation $o$ or hidden state $h^S$ (depending on the framework) to the meta-controller (Line 15). By following a greedy $\epsilon$ fashion, a subgoal will be selected from the meta-controller if it is a random number greater than $\epsilon$. Otherwise, a random subgoal will be selected (Algorithm \ref{alg:greedy}). The sub-controller is taught to reach the subgoal; when the subgoal is reached, a new subgoal will be selected. The process is repeated until the final goal is obtained. Intrinsic reward is evaluated by the critic module and is stored in $M^S$ (Line 19) for updating the sub-controller. Meanwhile, the extrinsic reward is directly received from the environment and is stored in $M^M$ for updating the meta-controller (Line 24) . Updating controllers at Line 20 and 21 is described in Section \ref{sec:learn} and is summarized in Algorithm \ref{alg:meta} and Algorithm \ref{alg:sub}.
\begin{algorithm}[ht]
\begin{algorithmic}[1]
\REQUIRE
{
\STATE x: input of the Q network
\STATE h: internal hidden states 
\STATE ${\cal B}$: a set of outputs
\STATE $\epsilon$: exploration rate
\STATE Q network and recurrent function $f$
}
\ENSURE
\STATE $h \leftarrow f(x,h)$ \;
\IF{$random() < \epsilon$}
\STATE $o \leftarrow$ An element from the set of output ${\cal B}$\;
\ELSE
\STATE $o = \operatorname*{argmax}_{m \in {\cal B}} Q(x,m)$\;
\ENDIF
\STATE \textbf{Return} $o, h$\; 
\caption{$EPS\_GREEDY(x, h, {\cal B}, \epsilon, Q, f)$}
\label{alg:greedy}
\end{algorithmic}
\end{algorithm}
\begin{algorithm}[ht]
\begin{algorithmic}[1]
\REQUIRE
{
\STATE $M^M$: experience replay memory of meta-controller
}
\ENSURE
\STATE \textbf{Sample} a mini-batch of $\{ o, g, r^{ex}, o' \}$ from $M^M$ as the strategy explained at \ref{sec:sample}\;
\STATE \textbf{Update the network} by minimizing the loss function: \[L^M =  \mathbb{E}_{(o,g,o',g',r^{ex}) \sim \cal {M}^M} \big[ y^M_i - Q^M(o,g | \theta^M)\big]\] where 
\[y^M_i = r^{ex} + \gamma Q^{M'}(o', \operatorname*{argmax}_{g'} Q^{M}(o',g' | \theta^{M}) |\theta^{M'})\]\;
\STATE \textbf{Update the target network: } $\theta^{M'} \leftarrow \tau\theta^M + (1-\tau)\theta^{M'}$
\caption{$META\_UPDATE(M^M, Q^M, Q^{M'})$}
\label{alg:meta}
\end{algorithmic}
\end{algorithm}
\begin{algorithm}[ht]
\begin{algorithmic}[1]
\REQUIRE
{
\STATE $M^S$: experience replay memory of meta-controller
}
\STATE \textbf{Sample} a mini-batch of $\{ \{o,g\},a,r^{in},\{o',g'\} \}$ from $M^S$ as the strategy explained at \ref{sec:sample}\;
\STATE \textbf{Update the main network} by minimizing the loss function: \[ L^S =  \mathbb{E}_{(o,g,a,r^{in}) \sim \cal {M}^S} \big[ y^S_i - Q^S(\{o,g\},a | \theta^S)\big],\]  where 
\[y^S_i = r^{in} + \gamma Q^{S'}(\{o', g\}, \operatorname*{argmax}_{a'} Q^S (\{o', g\},a' | \theta^{S'}) | \theta^{S'})\] \;
\STATE \textbf{Update the target network:} $\theta^{S'} \leftarrow \tau\theta^S + (1-\tau)\theta^{S'}$
\caption{$SUB\_UPDATE(M^S, Q^S, Q^{S'})$}
\label{alg:sub}
\end{algorithmic}
\end{algorithm}
\section{Experiments}
\label{sec:exp}

In this section, we evaluate two versions of the hierarchical deep recurrent network algorithm. hDRQNv1 is the algorithm formed by framework 1, and hDRQNv2 is the algorithm formed by framework 2. We compare them with flat algorithms (DQN, DRQN) and the state-of-the-art hierarchical RL algorithm (hDQN). The comparisons are performed on three domains. The domain of multiple goals in a gridworld is used to evaluate many aspects of the proposed algorithms. Meanwhile, the harder domain called multiple goals in four-rooms is used to benchmark the proposed algorithm. Finally, the most challenging game in ATARI 2600 \citep{mnih2013playing} called Montezuma's Revenge, is used to confirm the efficiency of our proposed framework.

\subsection{Implementation Details}
\label{sec:settings}

We use Tensorflow \citep{tensorflow2015-whitepaper} to implement our algorithms. The settings for each domain are different, but they have some commonalities as follows. For a hDRQNv1 algorithm, the inputs to the meta-controller and sub-controller are an image of size $44 \times 44 \times 3$ (a color image). The input image is resized from an observation which is observed around the agent (either $3 \times 3$ unit or $5 \times 5$ unit). The image feature of 256 values extracted through four CNNs and ReLUs is put into a LSTM layer of 256 states to generate 256 output values, and an internal hidden state of 256 values is also constructed. For a hDRQNv2 algorithm, a hidden state of 256 values is put into the network of the meta-controller. The state is passed through four fully connected layers and ReLU layers instead of four CONV layers. The output is a feature of 256 values. The algorithm uses ADAM \citep{kingma2014adam} for learning the neural network parameters with the learning rate $0.001$ for both the meta-controller network and sub-controller. For updating the target network, a $\tau$ value of $0.001$ is applied. The algorithm uses a discount factor of $0.99$. The capacity of $M^S$ and $M^M$ is $1E5$ and $5E4$, respectively.

\subsection{Multiple Goals in a Gridworld}
\label{task1}

\begin{figure}[htbp]
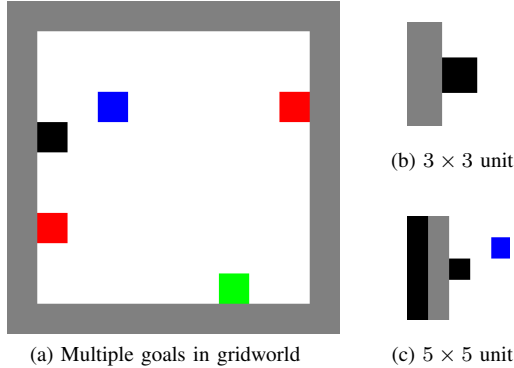

    \centering
    \begin{tabular}{cc}
    \adjustbox{valign=b}{
        \subfloat[Multiple goals in gridworld \label{fig:multiple_goal_gridworld_a}]{
          \includegraphics[width=.5\linewidth]{Fig5a.pdf}}}
    &
    \adjustbox{valign=b}{
    \begin{tabular}{@{}c@{}}
        \subfloat[$3 \times 3$ unit\label{fig:multiple_goal_gridworld_b}]{
            \includegraphics[width=.2\linewidth]{Fig5b.pdf}} \\
        \subfloat[$5 \times 5$ unit\label{fig:multiple_goal_gridworld_c}]{
              \includegraphics[width=.2\linewidth]{Fig5c.pdf}}
        \end{tabular}}
    \end{tabular}
    \caption{Multiple goals in gridworld}\label{fig:multiple_goal_gridworld}
\end{figure}

The domain of multiple goals in a gridworld is a simple form of multiple goals in four-rooms which is described in Section \ref{sec:extrinsic}. In this domain, a gridworld map of size $11 \times 11$ unit is instead of four-rooms map. At each time step, the agent only observes a part of the surrounding environment, either $3 \times 3$ unit (Fig. \ref{fig:multiple_goal_gridworld_b}) or $5 \times 5$ unit (Fig. \ref{fig:multiple_goal_gridworld_c}). The agent is allowed to choose one of four actions (top, left, right, bottom) that are deterministic. The agent can not move if the action leads it to the wall. The rewards for the agent are defined as follows. If the agent hits an obstacle, it will receive a penalty of minus one. If the agent reaches two goals in proper order, it will receive a reward of one for each goal. Otherwise, it only receives $0.01$. For applying to an hDRQN algorithm, an intrinsic reward function and an extrinsic reward function are defined as follows    
\begin{equation}
r^{in} = \begin{cases}
1 & \text{obtain the goal}\\
-1 & \text{hit the obstacle}
\end{cases}
\label{eq:intrinsic}
\end{equation}
and
\begin{equation} 
r^{ex} = \begin{cases}
1 & \text{reach goals in order}\\
0.01 & \text{otherwise}
\end{cases}
\label{eq:extrinsic}
\end{equation}

\begin{figure*}[htbp]
    \centering
    \begin{tabular}{ccc}
    \begin{tabular}{@{}c@{}}
        \subfloat[hDRQNv1: reward\label{fig:selected_transition_a}]{
            \includegraphics[width=.25\linewidth]{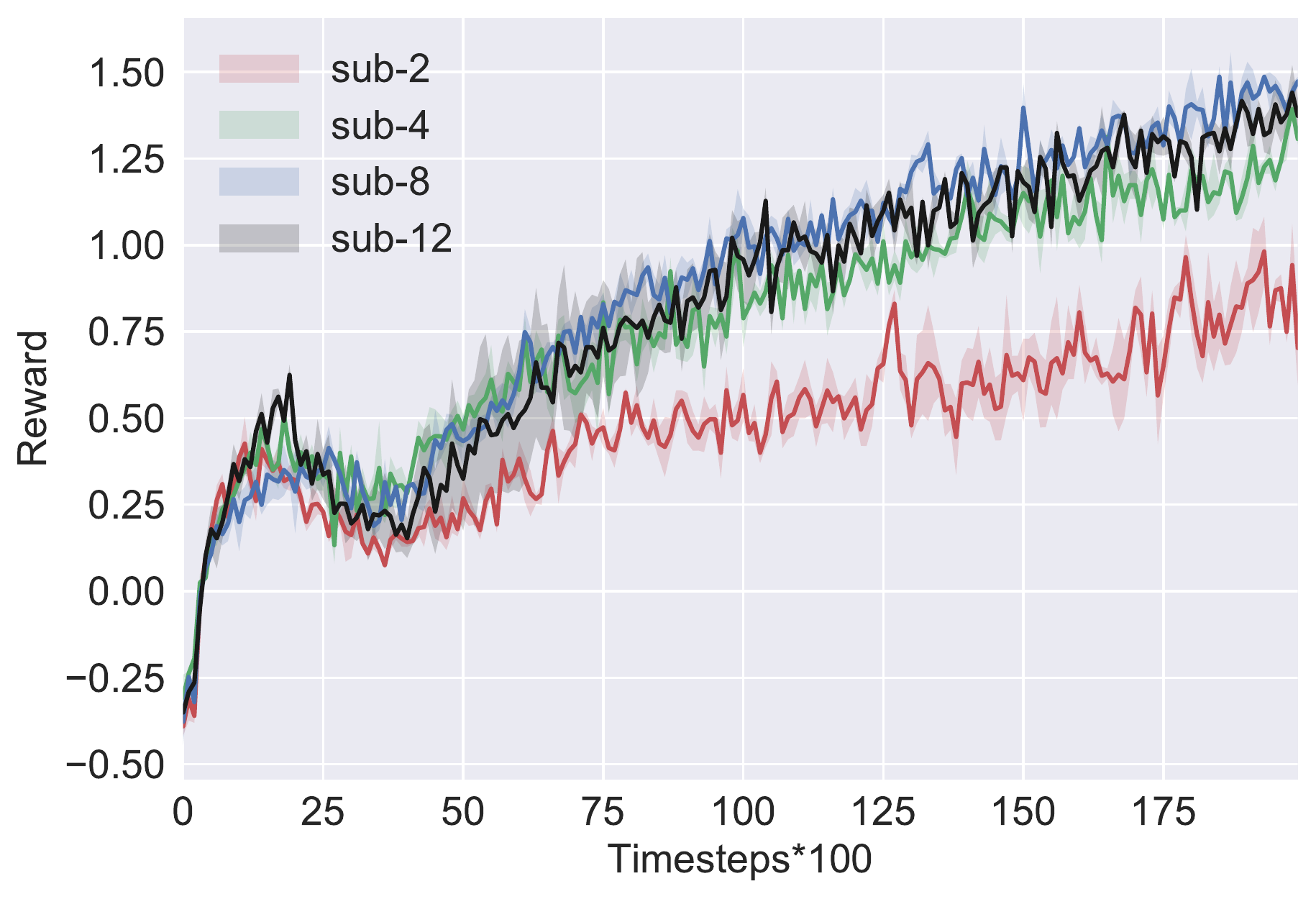}} \\
        \subfloat[hDRQNv1: intrinsic\label{fig:selected_transition_b}]{
              \includegraphics[width=.25\linewidth]{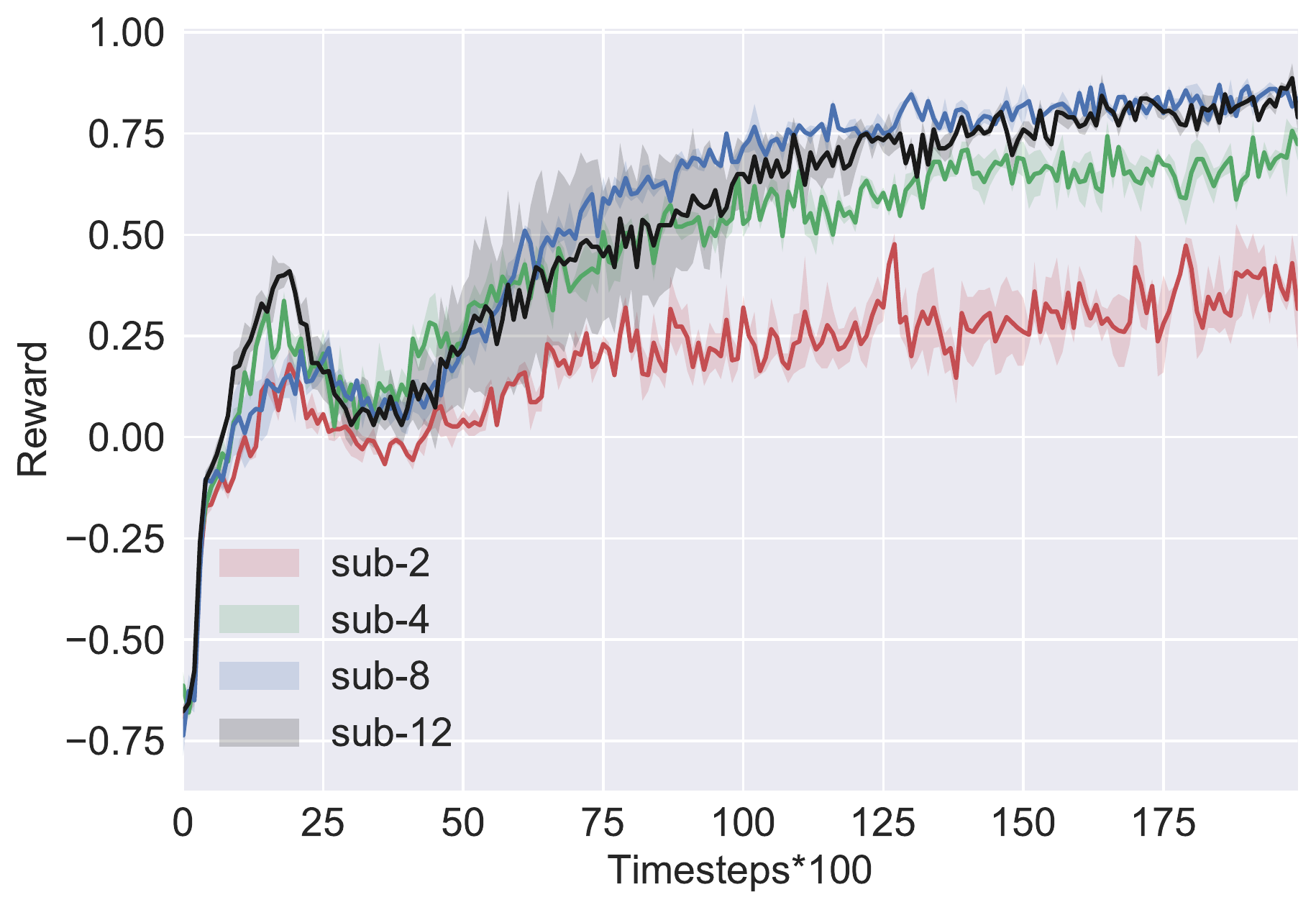}} \\
        \subfloat[hDRQNv1: extrinsic\label{fig:selected_transition_c}]{
              \includegraphics[width=.25\linewidth]{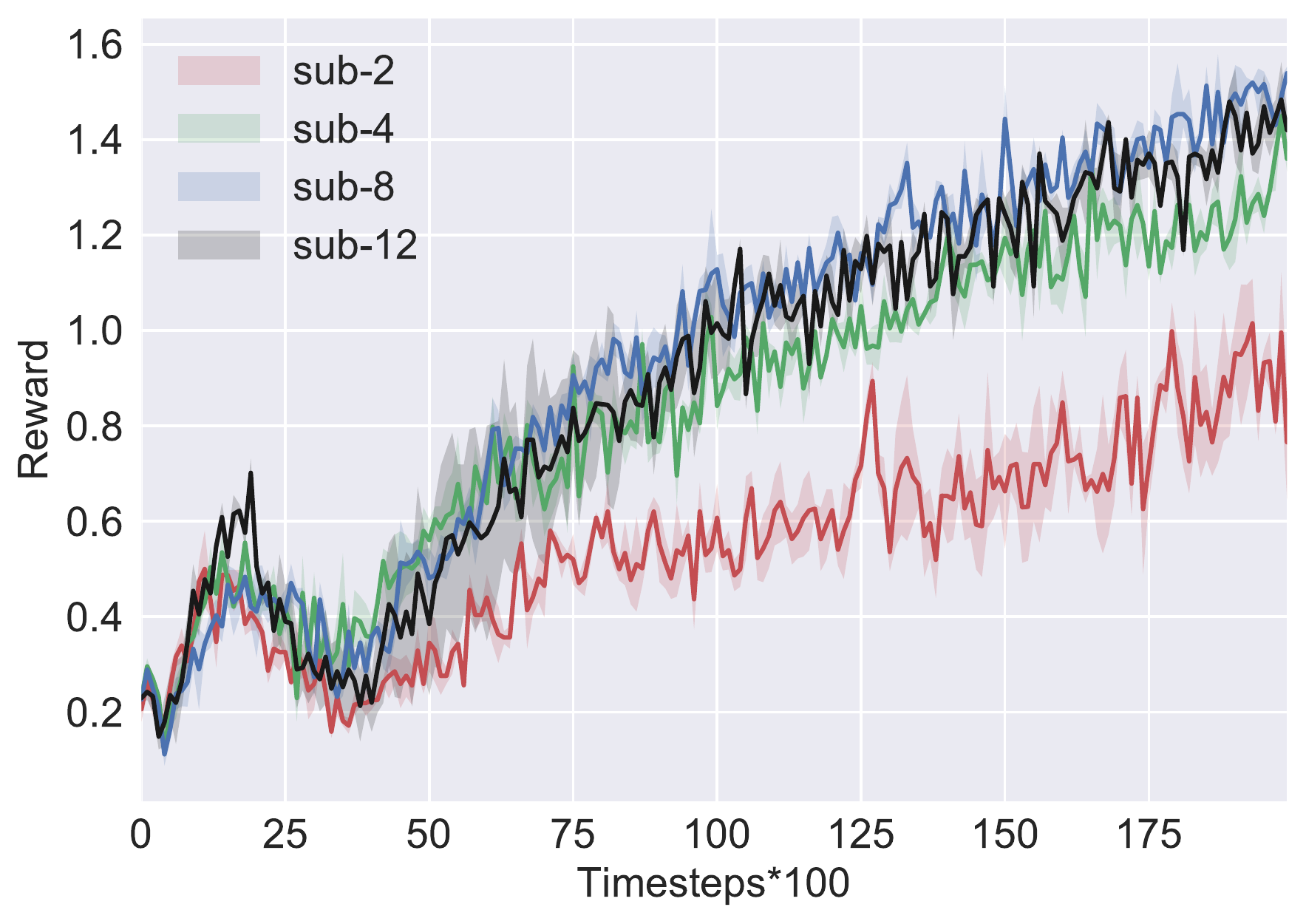}} \\
        \subfloat[hDRQNv1: steps\label{fig:selected_transition_d}]{
              \includegraphics[width=.25\linewidth]{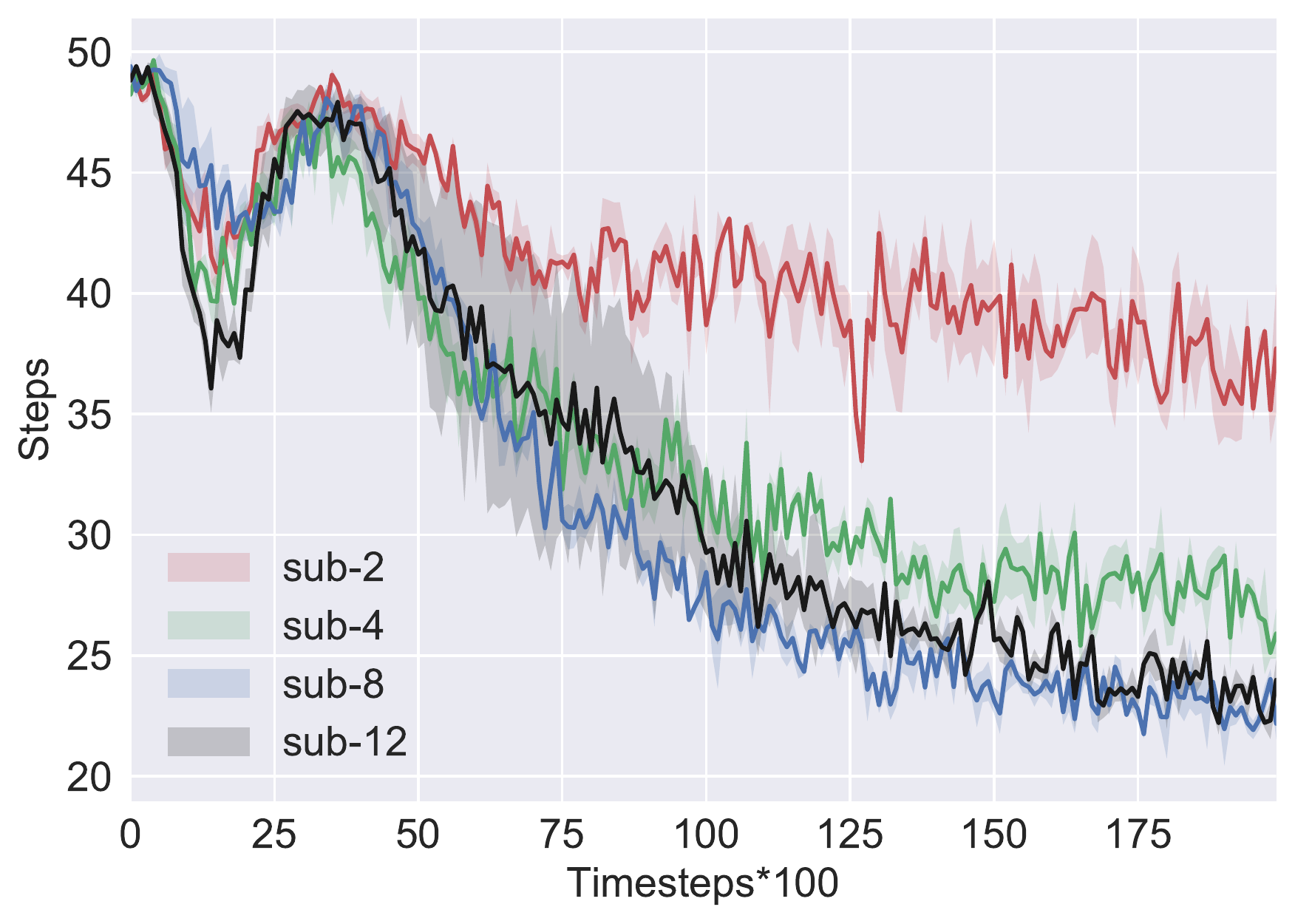}}
        \end{tabular}
    &
    \begin{tabular}{@{}c@{}}
        \subfloat[hDRQNv2: reward\label{fig:selected_transition_e}]{
            \includegraphics[width=.25\linewidth]{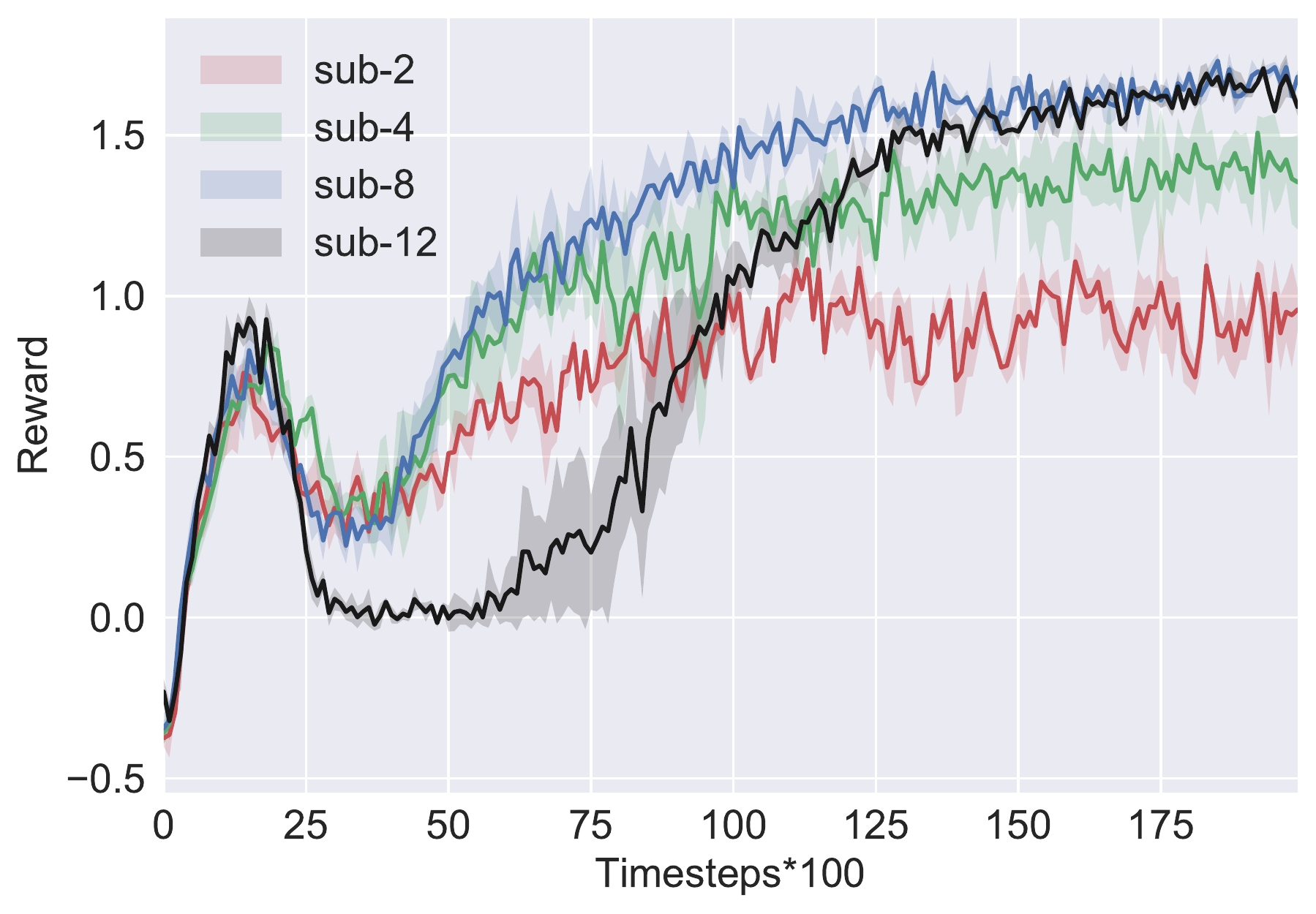}} \\
        \subfloat[hDRQNv2: intrinsic\label{fig:selected_transition_f}]{
              \includegraphics[width=.25\linewidth]{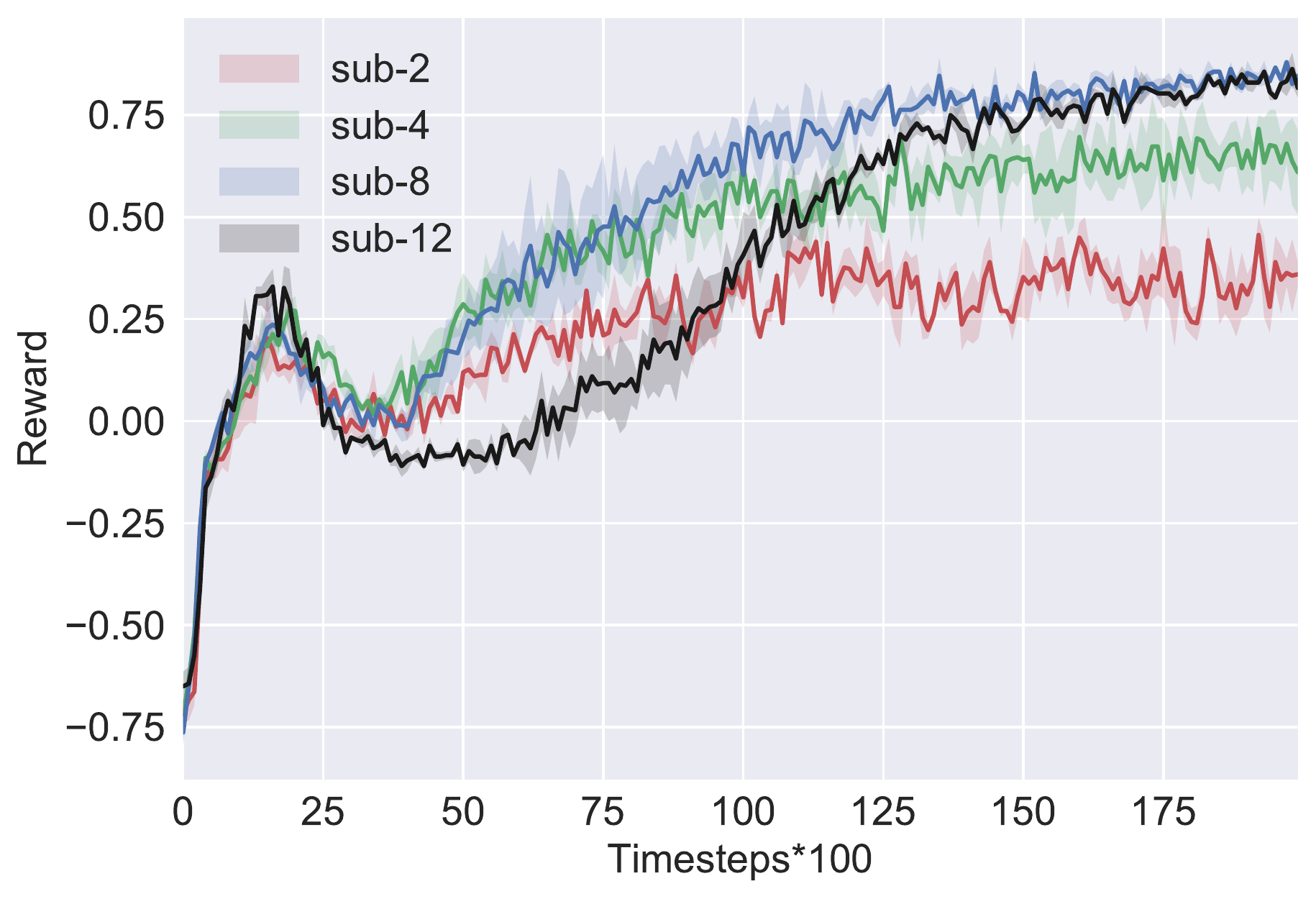}} \\
        \subfloat[hDRQNv2: extrinsic\label{fig:selected_transition_g}]{
              \includegraphics[width=.25\linewidth]{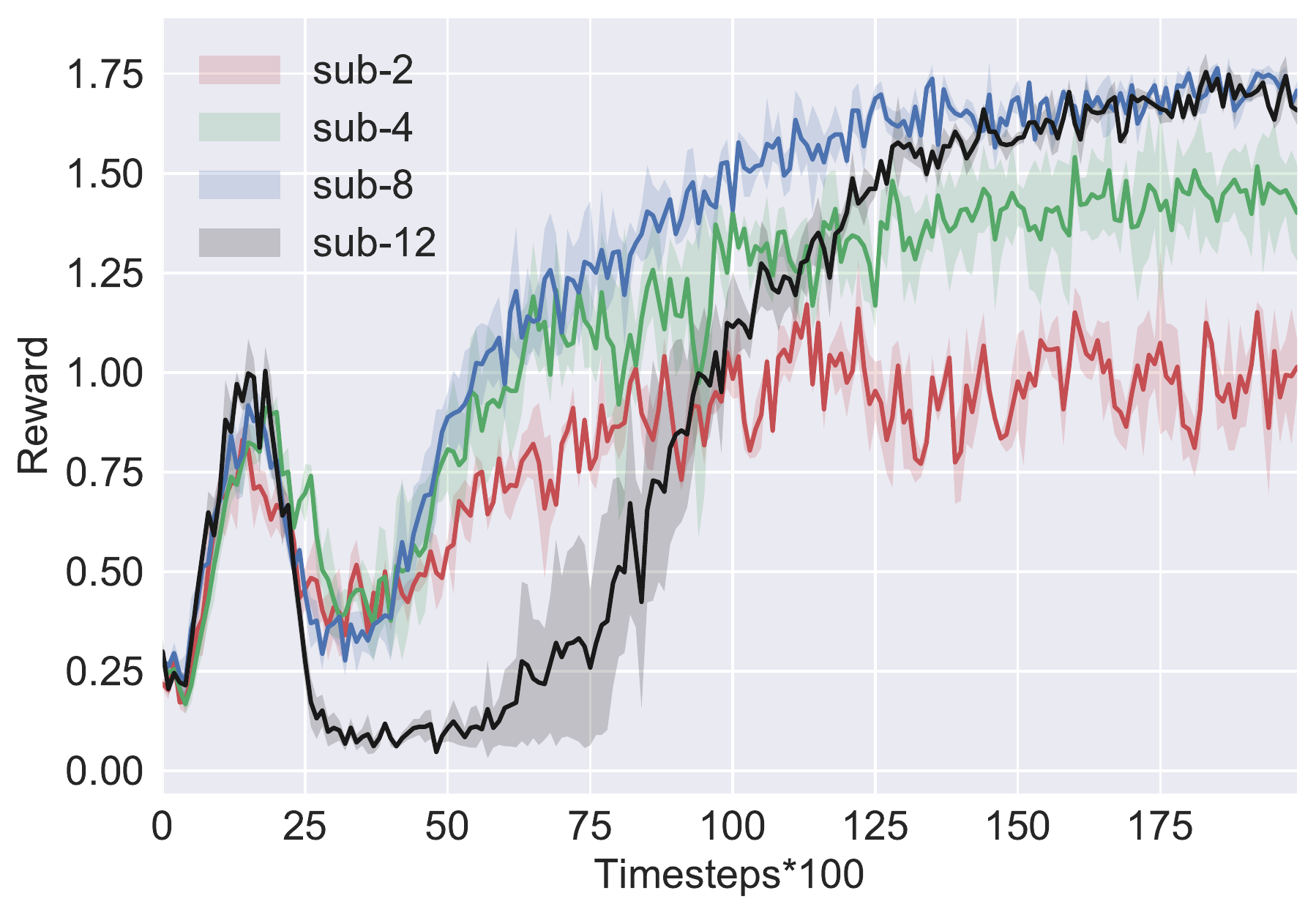}} \\
        \subfloat[hDRQNv2: steps\label{fig:selected_transition_h}]{
              \includegraphics[width=.25\linewidth]{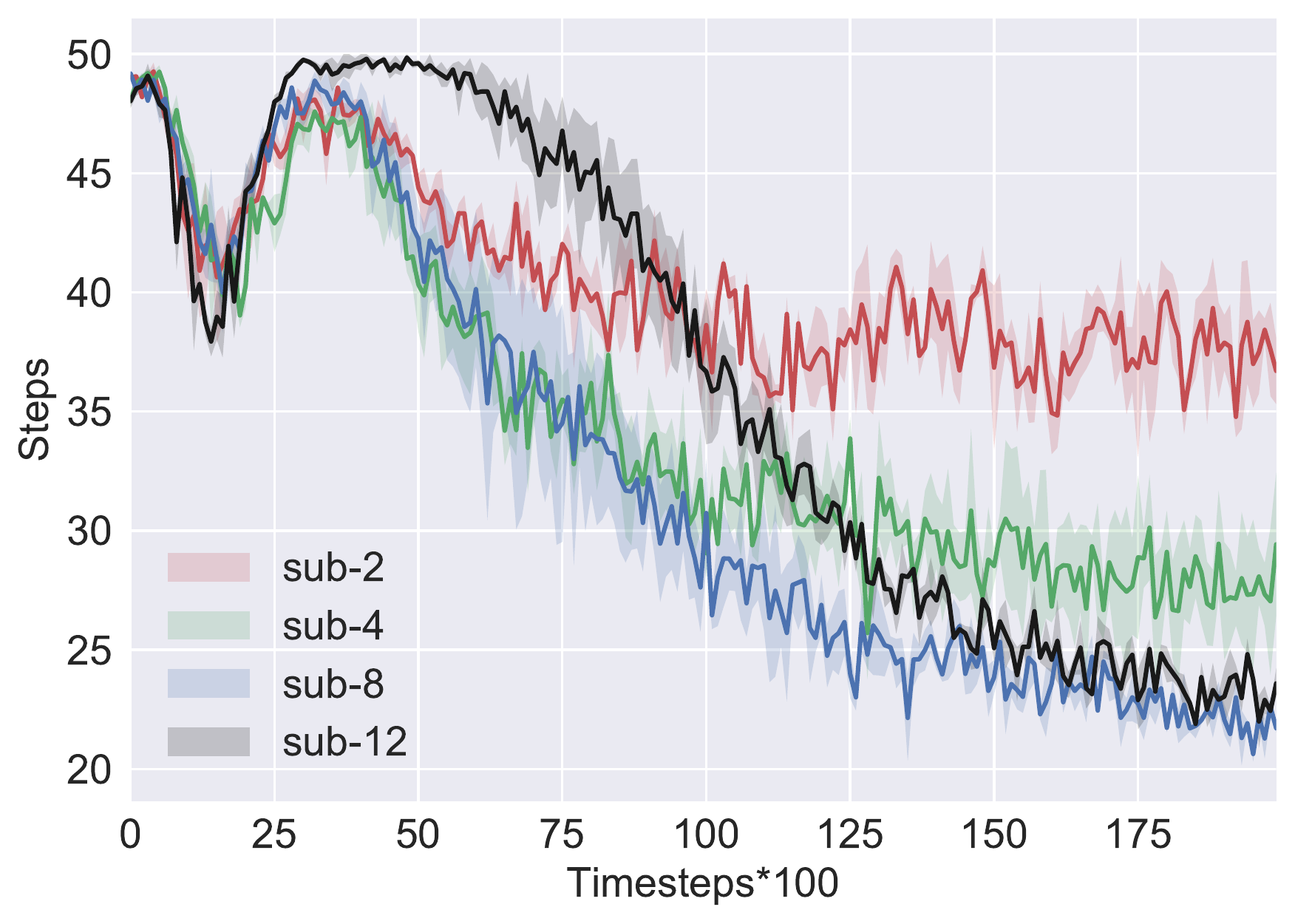}}
        \end{tabular}
    &
    \begin{tabular}{@{}c@{}}
        \subfloat[Reward\label{fig:selected_transition_i}]{
            \includegraphics[width=.25\linewidth]{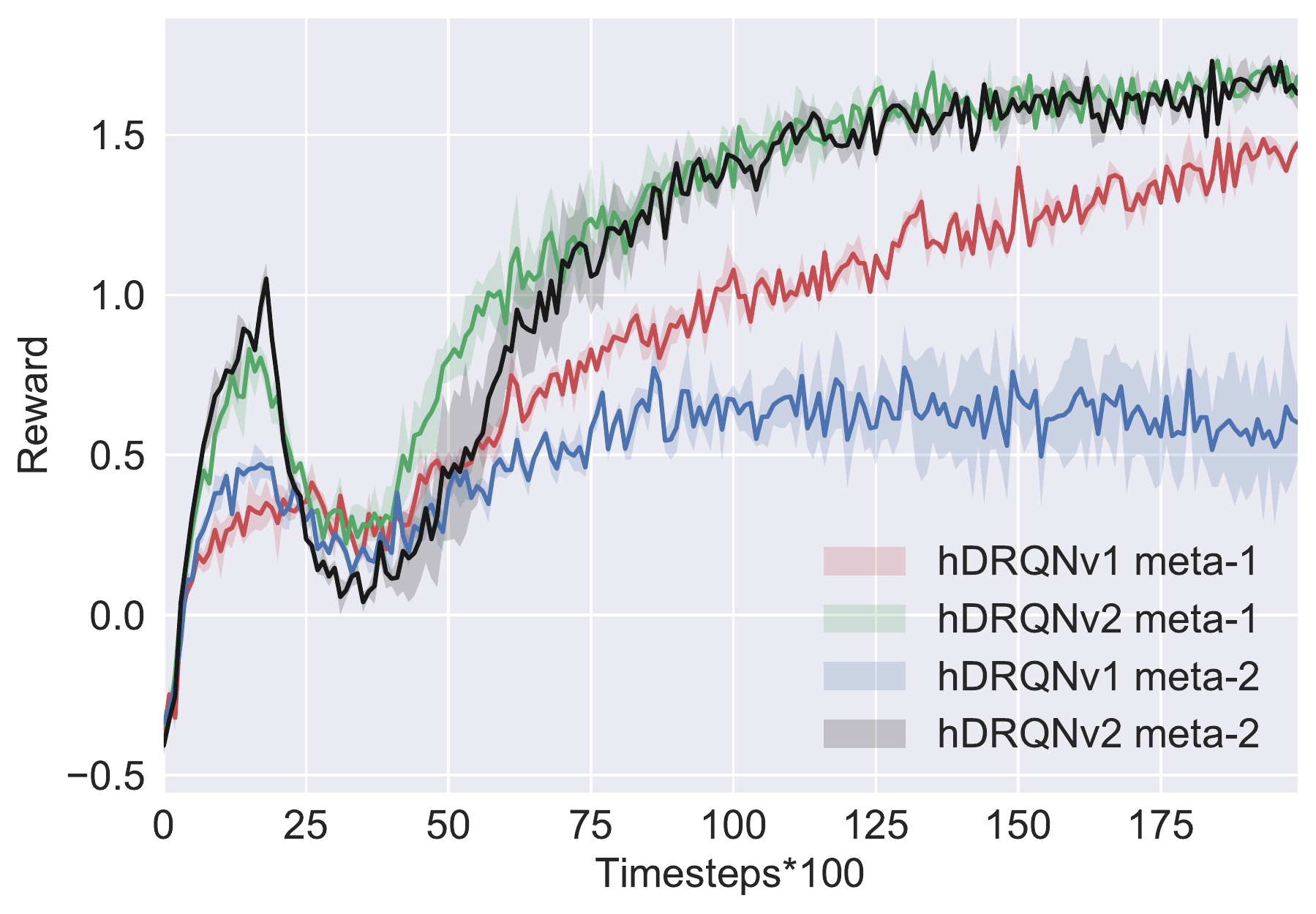}} \\
        \subfloat[Intrinsic\label{fig:selected_transition_j}]{
              \includegraphics[width=.25\linewidth]{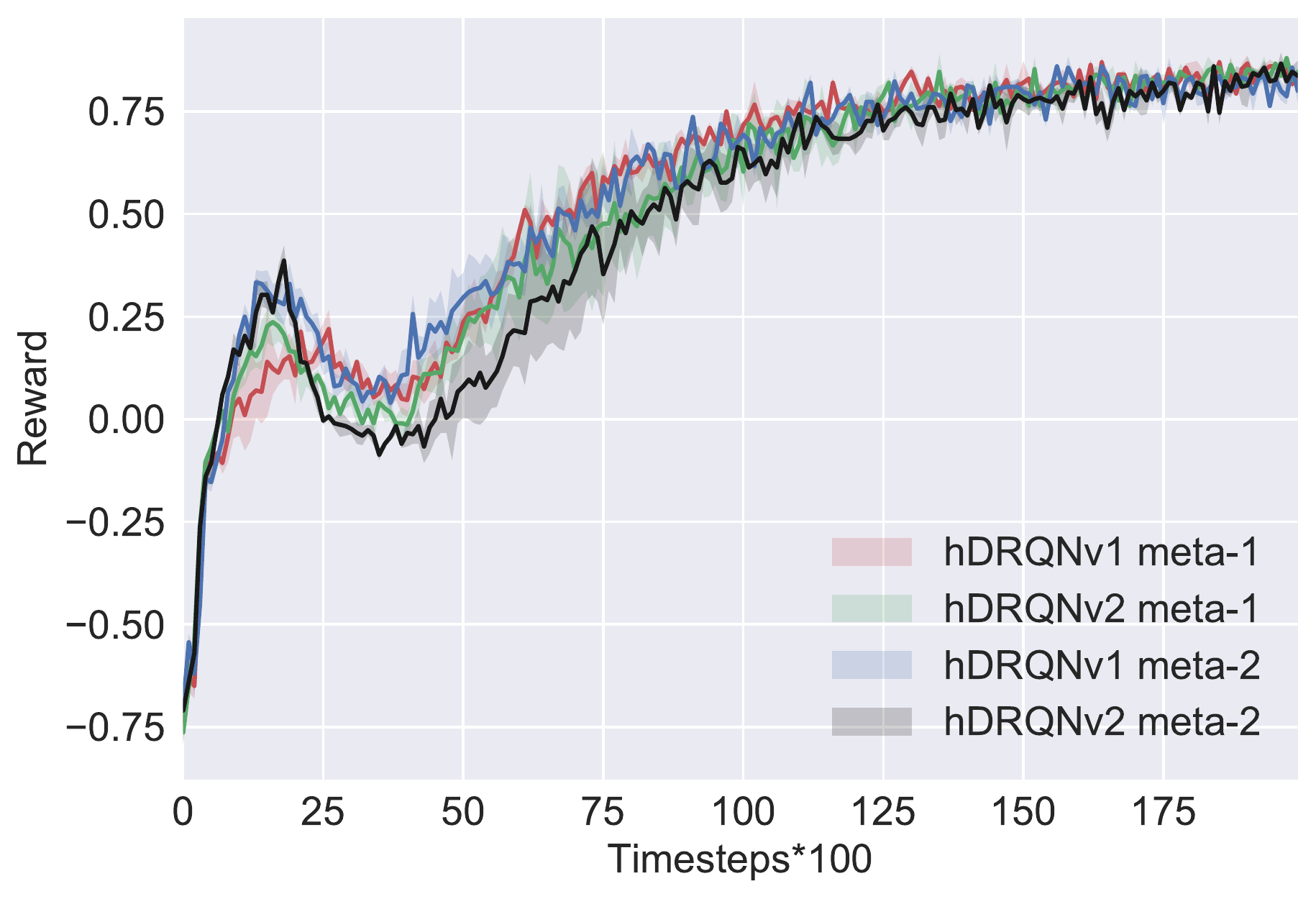}} \\
        \subfloat[Extrinsic\label{fig:selected_transition_k}]{
              \includegraphics[width=.25\linewidth]{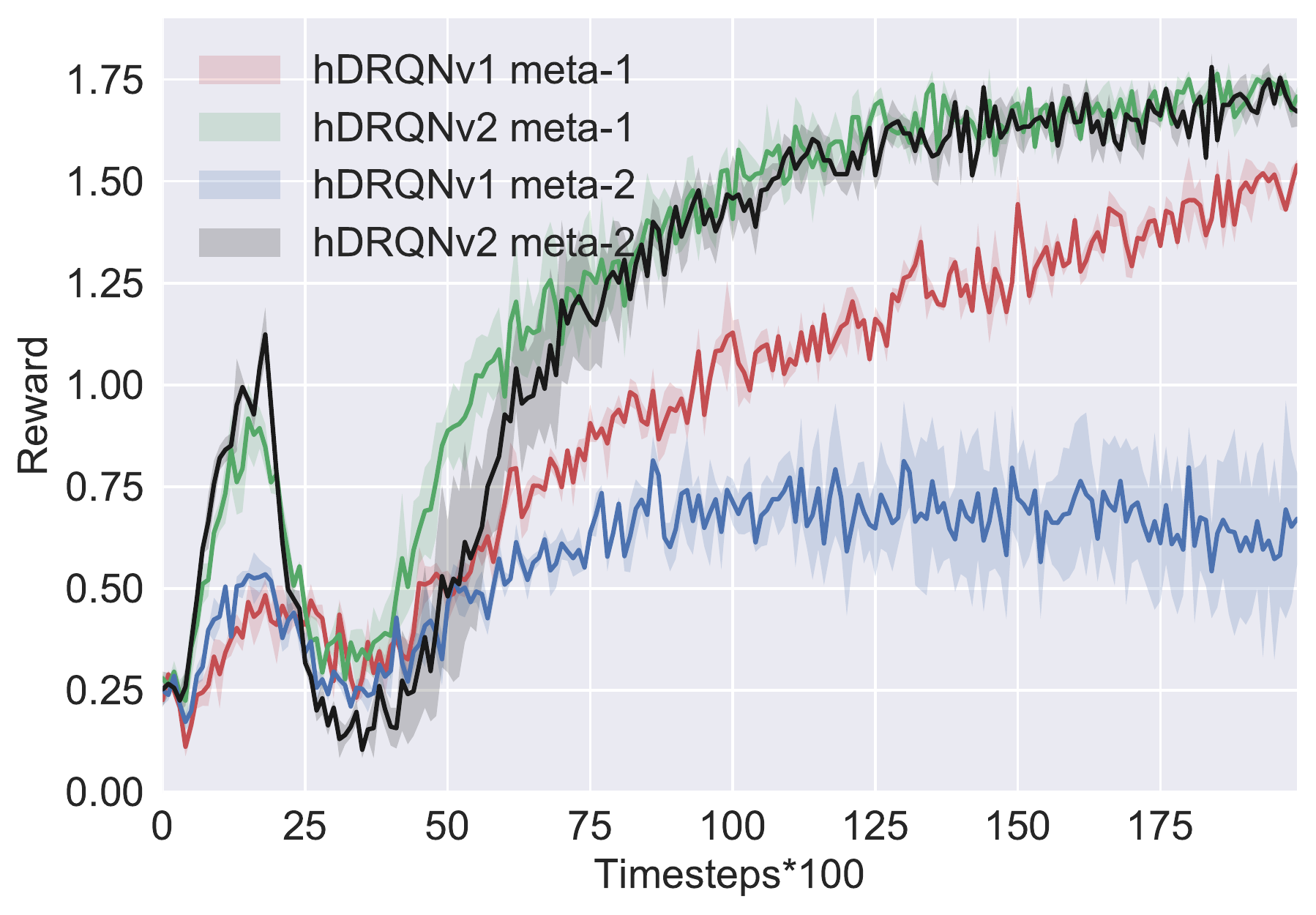}} \\
        \subfloat[Steps\label{fig:selected_transition_l}]{
              \includegraphics[width=.25\linewidth]{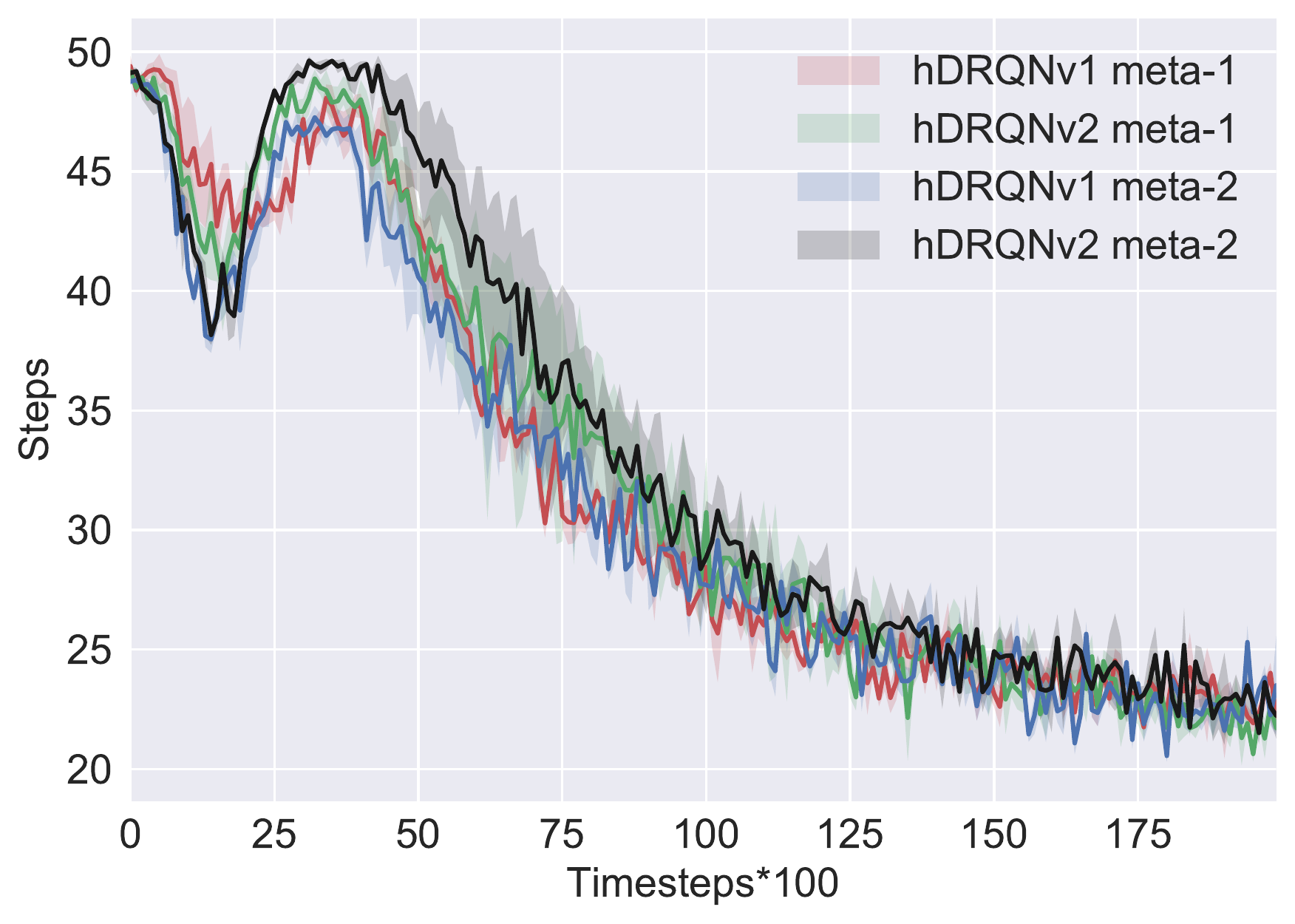}}
        \end{tabular}
    \end{tabular}
    \caption{Evaluation of different lengths of transitions. In the left-side figures, we use a fixed length of meta-transitions ($n^M = 1$). The figures on the right use a fixed length of sub-transitions ($n^S = 8$)}\label{fig:selected_transition}.
\end{figure*}

The first evaluation reported in Fig. \ref{fig:selected_transition} is a comparison of different lengths of selected transitions discussed in section \ref{sec:sample}). The agent in this evaluation can observe an area of $5 \times 5$ unit. We report the performance through three runs of 20000 episodes and each episode has 50 steps. The number of steps for each episode assures that the agent can explore any location on the map. In the figures on the left (hDRQNv1) and in the middle (hDRQNv2), we use a fixed length of meta-transitions ($n^M = 1$) and compare different lengths of sub-transitions. Otherwise, the figures on the right show the performance of the algorithm using a fixed length of sub-transitions ($n^S = 8$) and compare different lengths of meta-transitions. With a fixed length of meta-transition, the algorithm performs well with a long length of sub-transition ($n^S = 8$ or $n^S = 12$); the performance decreases when the length of sub-transitions is decreased. Intuitively, the RNN needs a sequence of transitions that is long enough to increase the probability that the agent will reach the subgoal within that sequence. Another observation is that, with $n^S = 8$ or $n^S = 12$, there is a little difference in performance. This is reasonable because only eight transitions are needed for the agent to reach the subgoals. For a fixed length of sub-transitions ($n^S = 8$), with a hDRQNv1 algorithm, the setting with $n^M = 2$ has low performance and high variance compared to the setting with $n^M = 1$. The reason is that while the sub-controller for two settings has the same performance (Fig. \ref{fig:selected_transition_f}) the meta-controller with $n^M = 1$ performs better than the meta-controller with $n^M = 2$. Meanwhile, with a hDRQNv2 algorithm, the performance is the same at both settings $n^M = 1$ and $n^M = 2$. This means that the hidden state from the sub-controller is a better input to determine the subgoal rather than a raw observation as it causes the algorithm to not depend on the length of the meta-transition. The number of steps to obtain two goals in order is around 22.

\begin{figure}[htbp]
\centering
\includegraphics[width=0.5\textwidth]{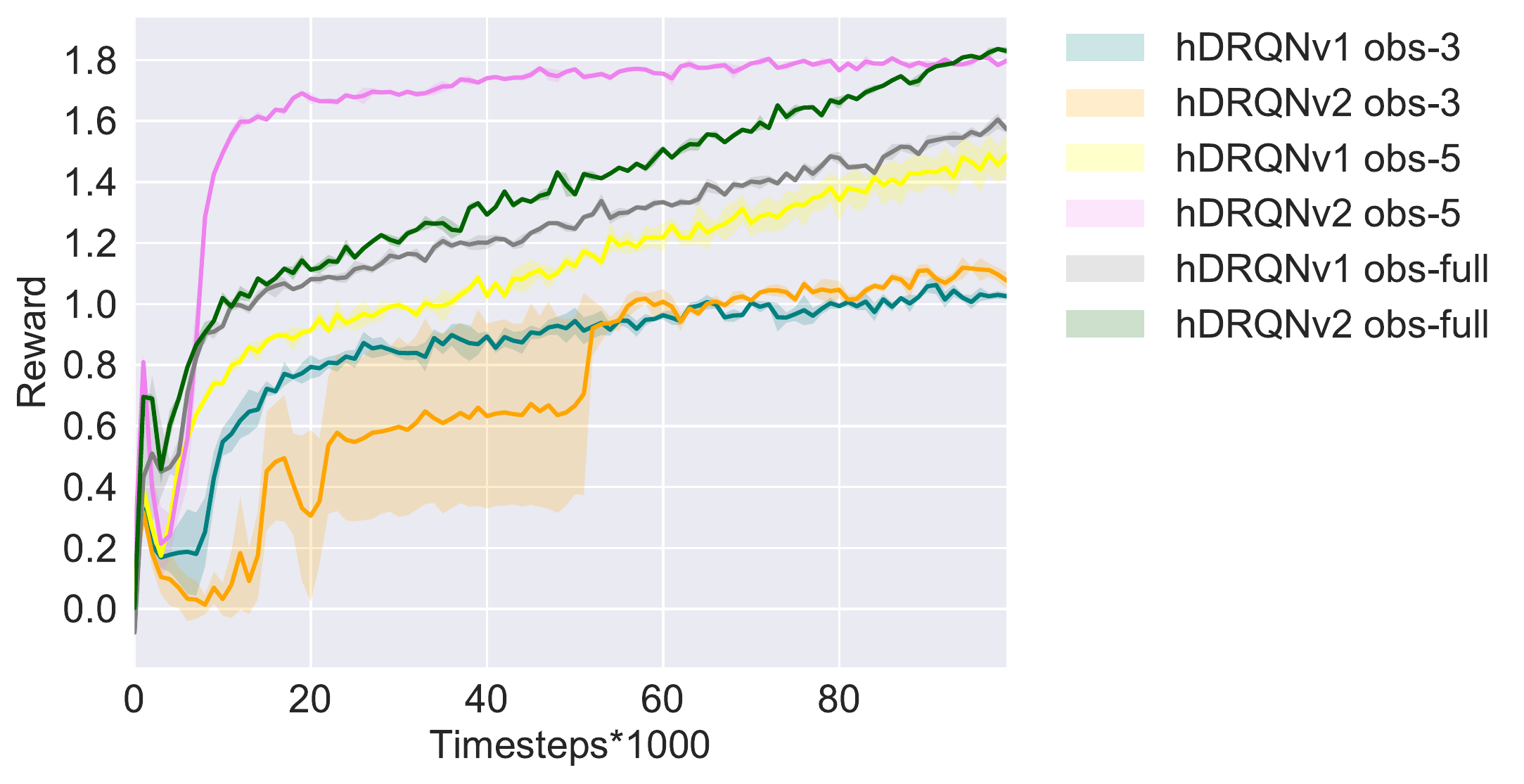}
\caption{Evaluation on different levels of observation}
\label{fig:evaluate_obs}
\end{figure}

The next evaluation is a comparison at different levels of observation. Fig. \ref{fig:evaluate_obs} shows the performance of hDRQN algorithms with a $3 \times 3$ observable agent compared with a $5 \times 5$ observable agent and a fully observable agent. It is clear that a fully observable agent can have more information around it than a $5 \times 5$ observable agent and a $3 \times 3$ observable agent; thus, the agent with a larger observation area can quickly explore and localize the environment completely. As a result, the performance of the agent with a larger observation area is better than the agent with a smaller observing ability. From the figure, the performance of a $5 \times 5$ observable agent using hDRQNv2 seems to converge faster than a fully observable agent. However, the performance of the fully observable agent surpasses the performance of $5 \times 5$ observable agent at the end.

\begin{figure}[htbp]
    \centering
    \begin{tabular}{c}
    \begin{tabular}{@{}c@{}}
        \subfloat[Two goals in gridworld\label{fig:compare_gridworld_a}]{
            \includegraphics[width=1.0\linewidth]{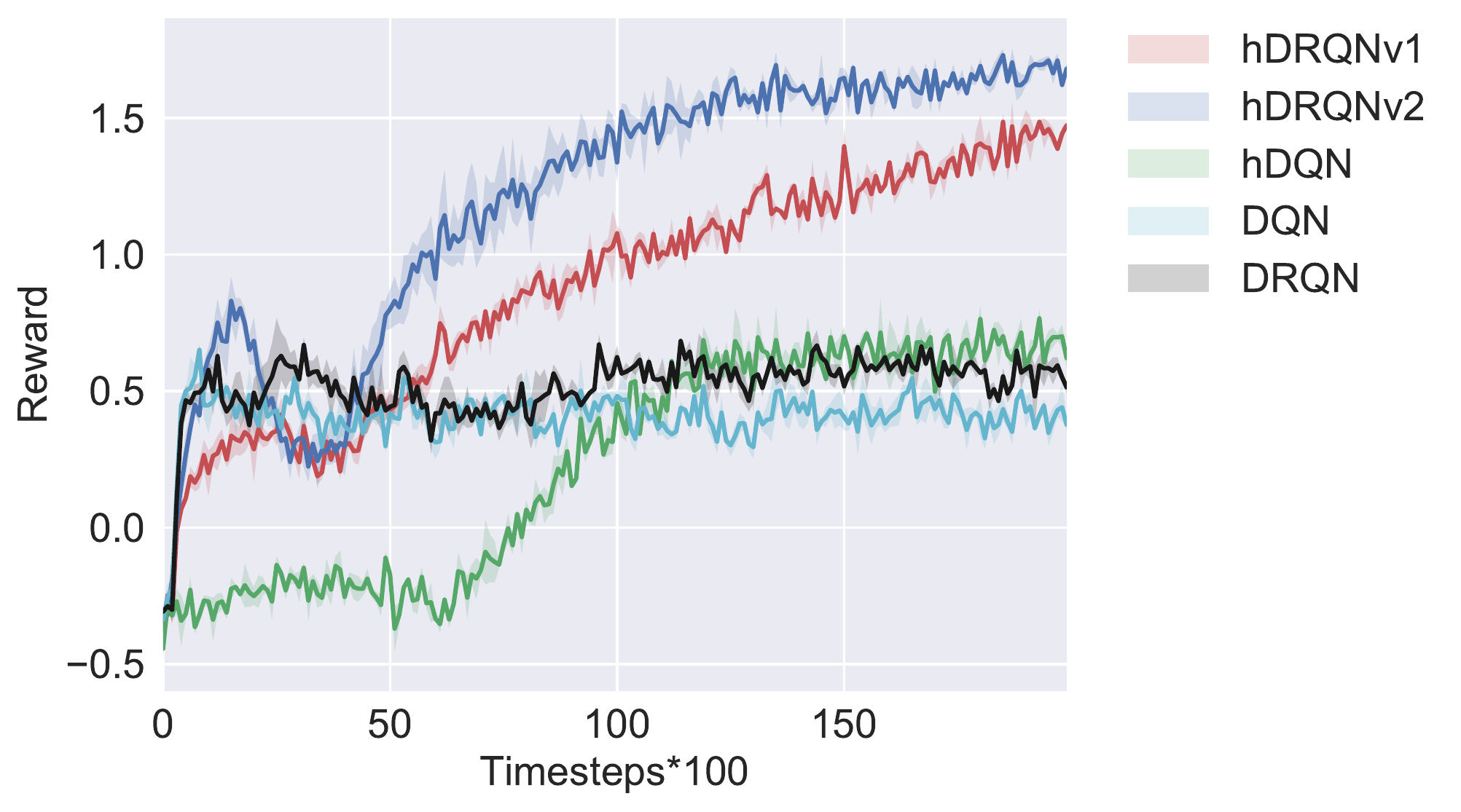}}\\
        \subfloat[Three goals in gridworld\label{fig:compare_gridworld_b}]{
            \includegraphics[width=1.0\linewidth]{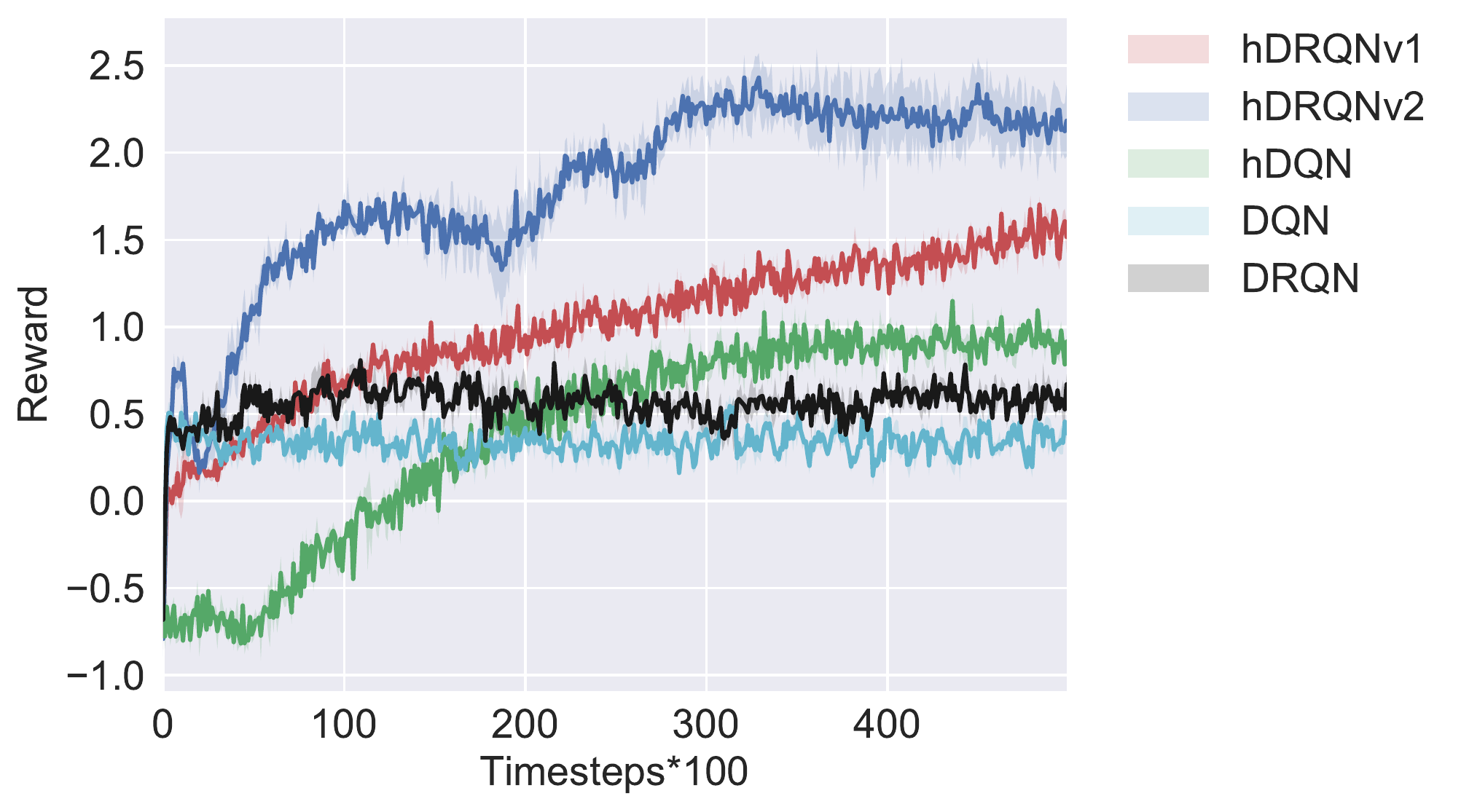}}
        \end{tabular}
    \end{tabular}
    \caption{Comparing hDRQN algorithms with some baseline algorithms}\label{fig:compare_gridworld}
\end{figure}

In the last evaluation of this domain, we compare the performance of the proposed algorithms with the well-known algorithms DQN, DRQN, and hDQN \citep{kulkarni2016hierarchical}. All algorithms assume that the agent only observes an area of $5 \time 5$ units around it. The results are shown in Fig. \ref{fig:compare_gridworld}. For both domains of two goals and three goals, hDRQN algorithms outperform other algorithms and hDRQNv2 has the best performance. The hDQN algorithm, which can operate in a hierarchical domain, is better than flat algorithms but not better than hDRQN algorithms. It might be that hDQN algorithm is only for fully observed domains and has poor performance in partially observable domains.

\subsection{Multiple Goals in a Four-rooms Domain}

\begin{figure}[htbp]
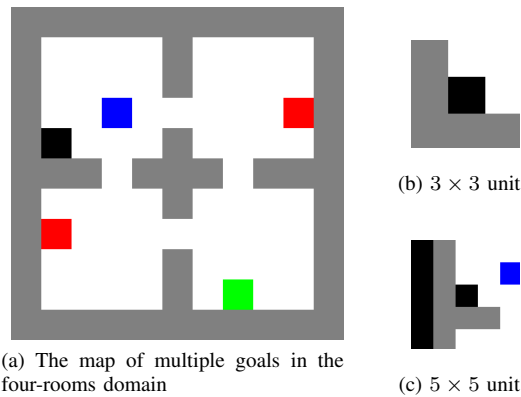

    \centering
    \begin{tabular}{cc}
    \adjustbox{valign=b}{
        \subfloat[The map of multiple goals in the four-rooms domain \label{fig:multiple_goal_fourroom_a}]{
          \includegraphics[width=.5\linewidth]{Fig9a.pdf}}}
    &
    \adjustbox{valign=b}{
    \begin{tabular}{@{}c@{}}
        \subfloat[$3 \times 3$ unit\label{fig:multiple_goal_fourroom_b}]{
            \includegraphics[width=.21\linewidth]{Fig9b.pdf}} \\
        \subfloat[$5 \times 5$ unit\label{fig:multiple_goal_fourroom_c}]{
              \includegraphics[width=.21\linewidth]{Fig9c.pdf}}
        \end{tabular}}
    \end{tabular}
    \caption{Multiple goals in the four-rooms domain}\label{fig:multiple_goal_fourroom}
\end{figure}

In this domain, we apply the multiple goals domain to a complex map called four-rooms (Fig. \ref{fig:multiple_goal_fourroom}). The dynamics of the environment in this domain is similar to that of the task in \ref{task1}. The agent in this domain must usually pass through hallways to obtain goals that are randomly located in four rooms. Originally, the four-rooms domain was an environment for testing a hierarchical reinforcement learning algorithm \citep{sutton1999between}. 
 
\begin{figure}[htbp]
    \centering
    \begin{tabular}{c}
    \begin{tabular}{@{}c@{}}
        \subfloat[Two goals in four-rooms\label{fig:compare_fourroom_a}]{
            \includegraphics[width=1.0\linewidth]{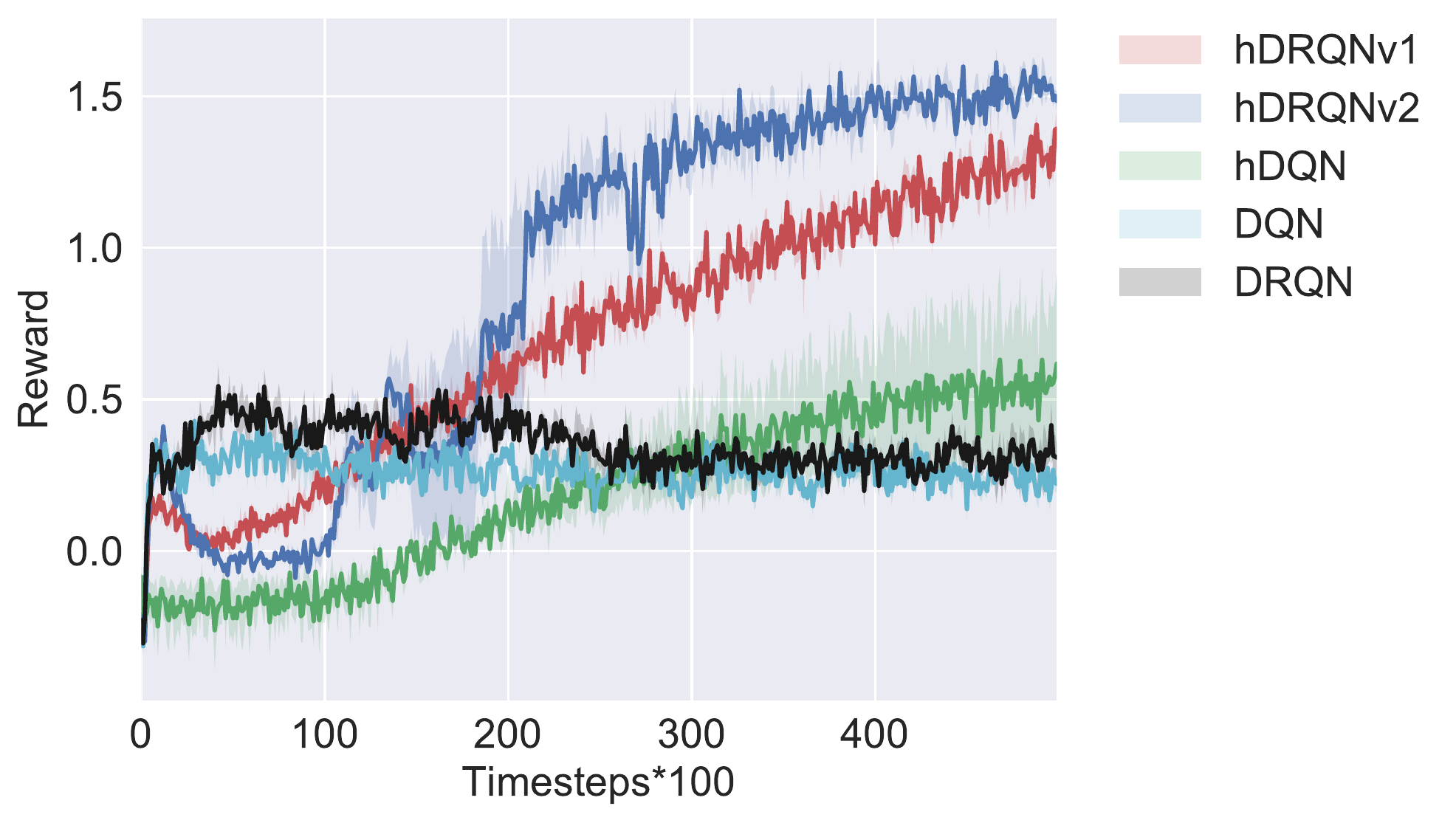}} \\
        \subfloat[Three goals in four-rooms\label{fig:compare_fourroom_b}]{
            \includegraphics[width=1.0\linewidth]{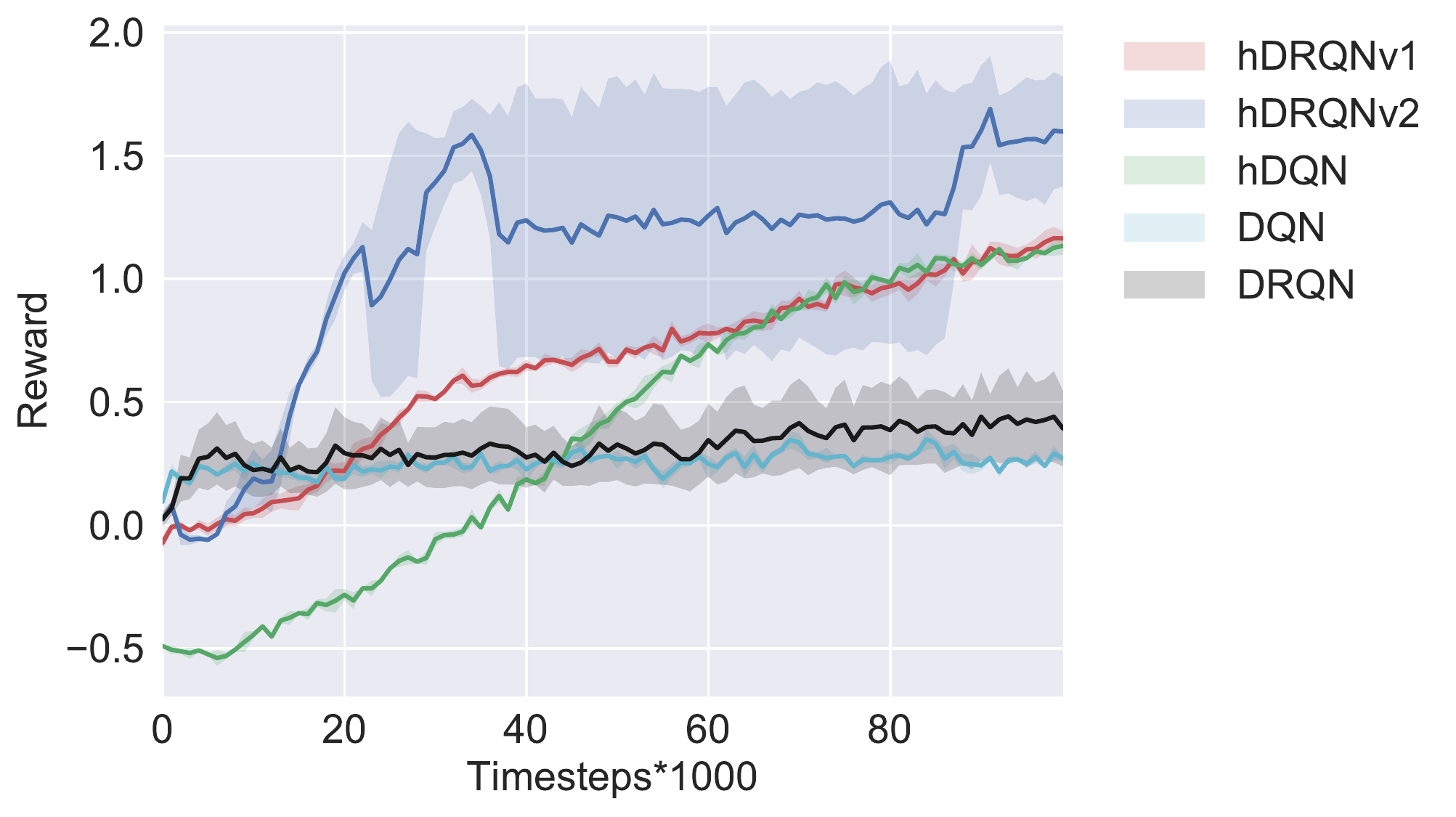}}
        \end{tabular}
    \end{tabular}
    \caption{Comparing hDRQN algorithms with some baseline algorithms}\label{fig:compare_fourroom}
\end{figure}

The performance is shown in Fig. \ref{fig:compare_fourroom_a} is averaged through three runs of 50000 episodes and each episode has 50 time steps. Meanwhile, the performance is shown in Fig. \ref{fig:compare_fourroom_b} is averaged through three runs of 100000 episodes, and each episode has 100 time steps. Similarly, the proposed algorithms outperform other algorithms, especially, the hDRQNv2 algorithm. 

\subsection{Montezuma's Revenge game in ATARI 2600}

\begin{figure}[htbp]
    \centering
    \begin{tabular}{cc}
    \adjustbox{valign=b}{
        \subfloat[The first screen of Montezuma's Revenge game\label{fig:monte_a}]{
          \includegraphics[width=.5\linewidth]{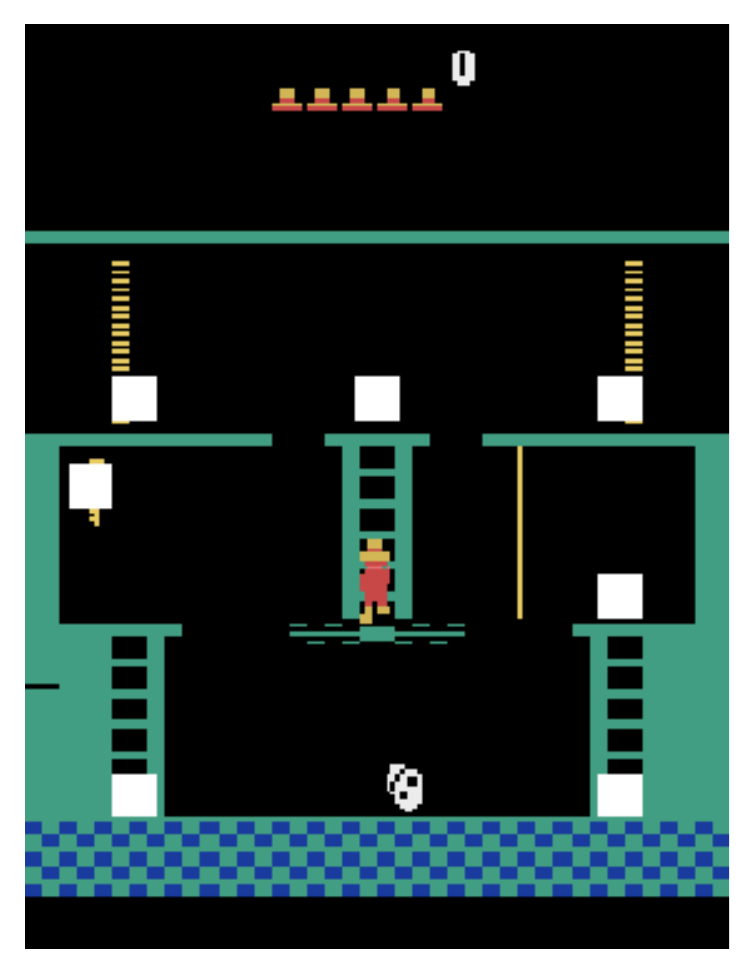}}}
    &
    \adjustbox{valign=b}{
    \begin{tabular}{@{}c@{}}
        \subfloat[Observable area\label{fig:monte_b}]{
            \includegraphics[width=.21\linewidth]{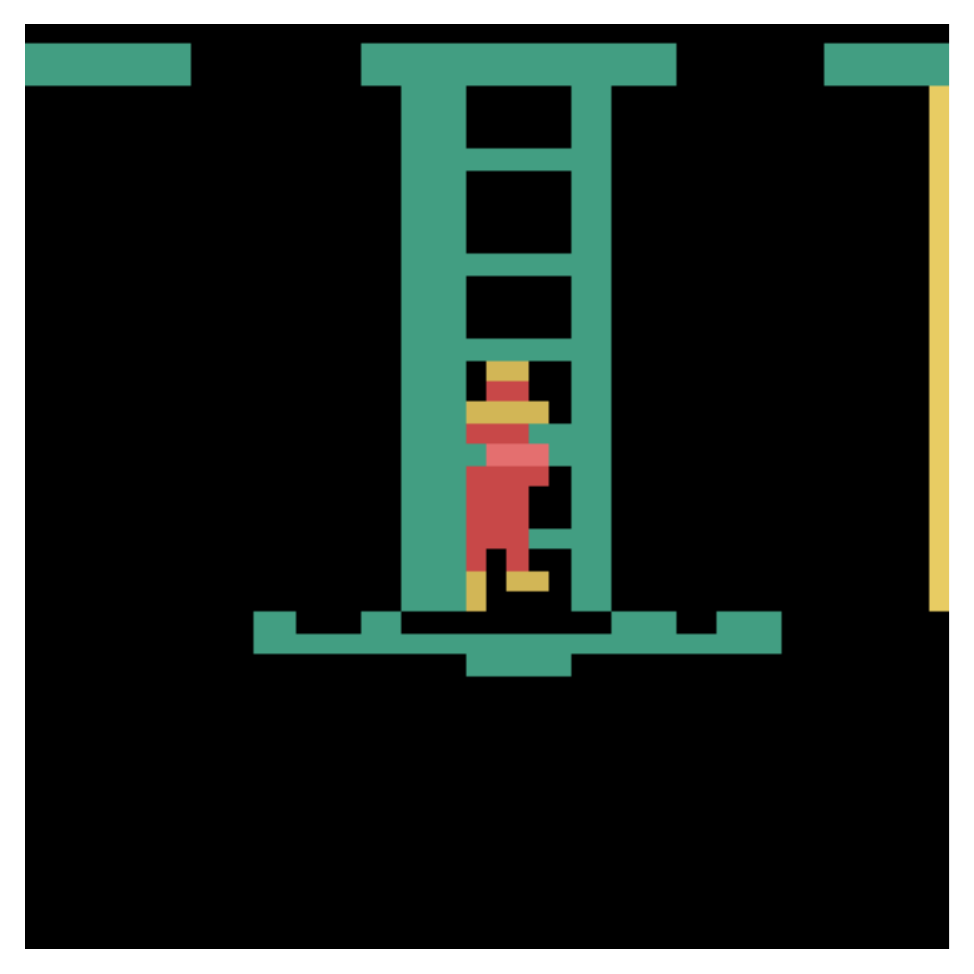}} \\
        \end{tabular}}
    \end{tabular}
    \caption{Montezuma's Revenge game in ATARI 2600}\label{fig:monte}
\end{figure}

Montezuma’s Revenge is the hardest game in ATARI 2600, where the DQN algorithm \citep{mnih2015human} can only achieve a score of zero. We use OpenAI Gym to simulate this domain \cite{openai}. The game is hard because the agent must execute a long sequence of actions until a state with non-zero reward (delayed reward) can be visited. In addition, in order to obtain a state with larger rewards, the agent needs to reach a special state in advance. This paper evaluates our proposed algorithms on the first screen of the game (Fig. \ref{fig:monte}). Particularly, the agent, which only observes a part of the environment (Fig. \ref{fig:monte_b}), needs to pass through doors (the yellow line in the top left and top right corners of the figure) to explore other screens. However, to pass through the doors, first, the agent needs to pick up the key on the left side of the screen. Thus, the agent must learn to navigate to the key's location and then navigate back to the door and open the next screens. The agent will earn 100 after it obtains the key and 300 after it reaches any door. Totally, the agent can receive 400 for this screen.

\begin{figure}[htbp]
\centering
\includegraphics[width=0.5\textwidth]{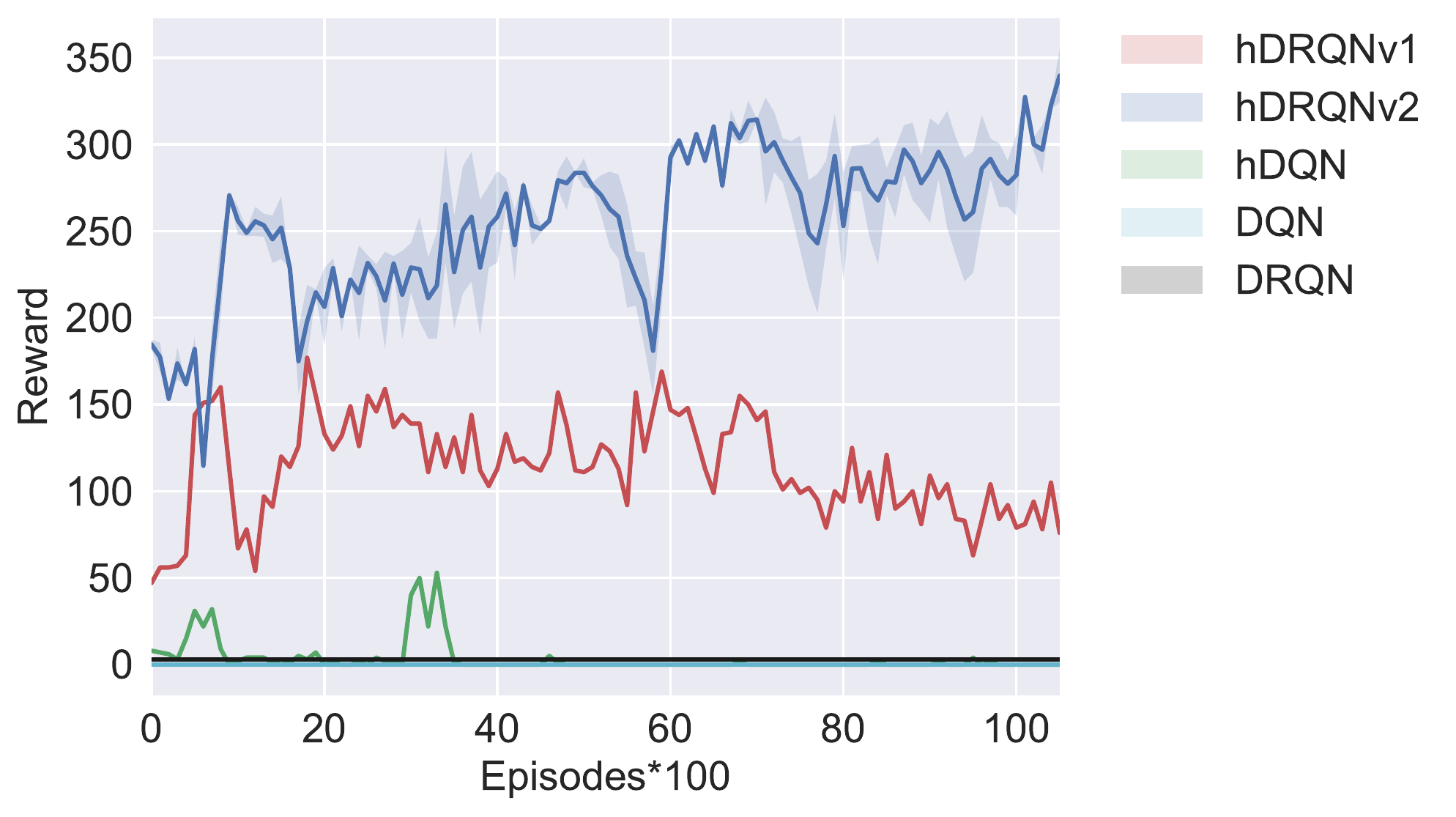}
\caption{Comparing hDRQN algorithms with some baseline algorithms on Montezuma's Revenge game}
\label{fig:compare_monte}
\end{figure}

\begin{figure}[htbp]
\centering
\includegraphics[width=0.4\textwidth]{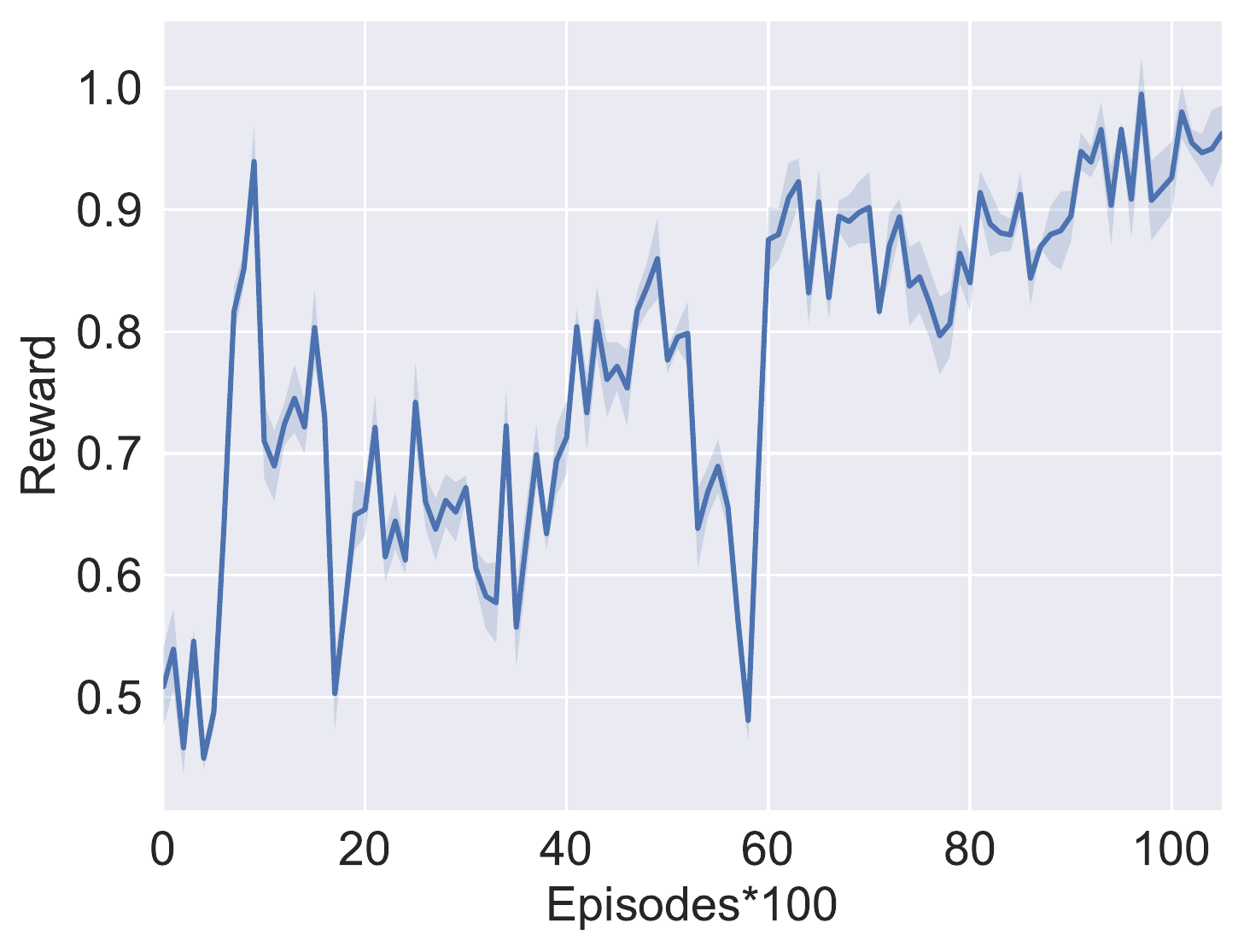}
\caption{Success ratio for reaching the goal ``key''}
\label{fig:reached_key}
\end{figure}

\begin{figure}[htbp]
\centering
\includegraphics[width=0.5\textwidth]{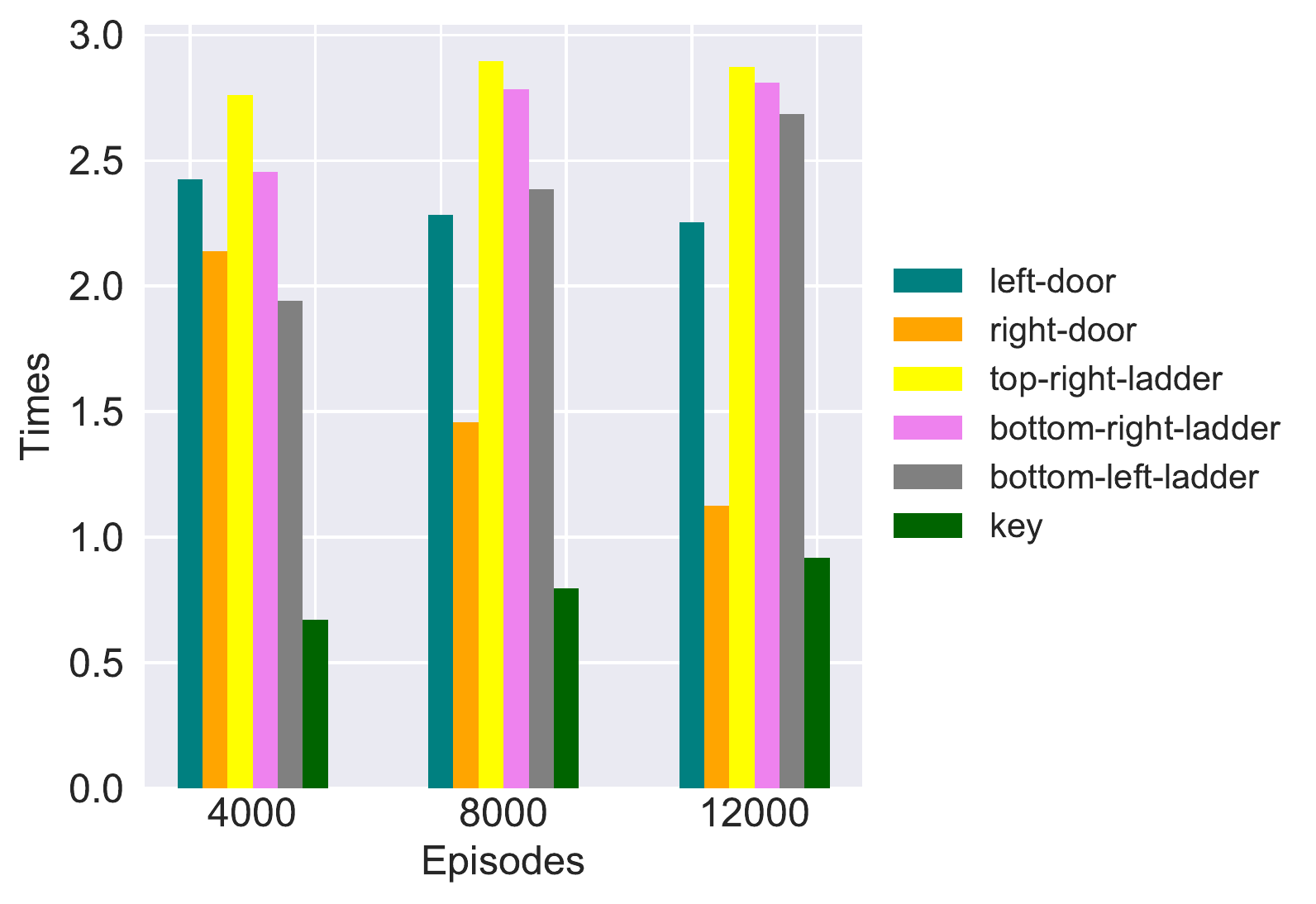}
\caption{Statistic about the number of times the agent visits subgoals}
\label{fig:reached_states}
\end{figure}

The intrinsic reward function is defined to motivate the agent to explore the whole environment. Particularly, the agent will be received an intrinsic value of 1 if it could reach a subgoal from other subgoals. The set of subgoals is pre-defined in Fig. \ref{fig:monte_a} (the white rectangles). In contrast to intrinsic reward function, the extrinsic reward function is defined as a reward value of 1 when the agent obtains the key or open the doors. Because learning the meta-controller and the sub-controllers simultaneously is highly complex and time-consuming, we separate the learning process into two phases. In the first phase, we learn the sub-controllers completely such that the agent can explore the whole environment by moving between subgoals. In the second phase, we learn the meta-controller and sub-controllers altogether. The architecture of the meta-controller and the sub-controllers is described in section \ref{sec:settings}. The length of sub-transition and meta-transition is $8$ and $16$ correspondingly. In this domain, the agent can observe an area of $70 \times 70$ pixels. Then, the observation area is resized to $44 \times 44$ to fit the input of a controller network. The performance of proposed algorithms compared to baseline algorithms is shown in Fig. \ref{fig:compare_monte}. DQN reported a score of zero which is similar to the result from \citep{mnih2015human}. DRQN which can perform well in a partially observable environment also achieves a score of zero because of the highly hierarchical complexity of the domain. Meanwhile, hDQN can achieve a high score on this domain. However, it cannot perform well in a partial observability setting. The performance of hDQN in a full observability setting can be found in the paper of Kulkarni \citep{kulkarni2016hierarchical}. Our proposed algorithms can adapt to the partial observability setting and hierarchical domains as well. The hDRQNv2 algorithm shows a better performance than hDRQNv1. It seems that the difference in the architecture of two frameworks (described in section \ref{sec:algo}) has affected their performance. Particularly, using internal states of a sub-controller as the input to the meta-controller can give more information for prediction than using only a raw observation. To evaluate the effectiveness of the two algorithms, we report the success ratio for reaching the goal ``key'' in Fig. \ref{fig:reached_key} and the number of time steps the agent explores each subgoal in Fig. \ref{fig:reached_states}. In Fig. \ref{fig:reached_key}, the agent using hDRQNv2 algorithm almost picks up the ``key'' at the end of the learning process. Moreover, Fig. \ref{fig:reached_states} shows that hDRQNv2 tends to explore more often on subgoals that are on the way reaching the ``key'' (E.g. top-right-ladder, bottom-right-ladder, and bottom-left-ladder) while exploring less often on other subgoals such as the left door and right door.

\section{Conclusion}
\label{sec:con}

We introduced a new hierarchical deep reinforcement learning algorithm that is a learning framework for both full observability (MDP) and partial observability (POMDP). The algorithm takes advantages of deep neural networks (DNN, CNN, LSTM) to produce hierarchical policies that can solve domains with a highly hierarchical nonlinearity. We showed that the framework has performed well when learning in hierarchical POMDP environments. Nevertheless, our approach contains some limitations as follows. First, our framework is built on two levels of a hierarchy which does not fit into the domain with multiple levels of hierarchy. Second, in order to simplify the learning problem in hierarchical POMDP, we assume that the set of subgoals is predefined and fixed because the problem of discovering a set of subgoals in POMDP is still a hard problem. In the future, we plan to improve our framework by tackling those problems. In addition, we can apply DRQN to multi-agent problems where the environment is partially observable and the task is hierarchical.

\section*{Acknowledgment}

The authors are grateful to the Basic Science Research Program through the National Research Foundation of Korea (NRF-2017R1D1A1B04036354).

\ifCLASSOPTIONcaptionsoff
  \newpage
\fi



\bibliographystyle{IEEEtran}      
\bibliography{bibfile}

\begin{thebibliography}{10}
\providecommand{\url}[1]{#1}
\csname url@samestyle\endcsname
\providecommand{\newblock}{\relax}
\providecommand{\bibinfo}[2]{#2}
\providecommand{\BIBentrySTDinterwordspacing}{\spaceskip=0pt\relax}
\providecommand{\BIBentryALTinterwordstretchfactor}{4}
\providecommand{\BIBentryALTinterwordspacing}{\spaceskip=\fontdimen2\font plus
\BIBentryALTinterwordstretchfactor\fontdimen3\font minus
  \fontdimen4\font\relax}
\providecommand{\BIBforeignlanguage}[2]{{%
\expandafter\ifx\csname l@#1\endcsname\relax
\typeout{** WARNING: IEEEtran.bst: No hyphenation pattern has been}%
\typeout{** loaded for the language `#1'. Using the pattern for}%
\typeout{** the default language instead.}%
\else
\language=\csname l@#1\endcsname
\fi
#2}}
\providecommand{\BIBdecl}{\relax}
\BIBdecl

\bibitem{Sutton:1998:IRL:551283}
R.~S. Sutton and A.~G. Barto, \emph{Introduction to Reinforcement Learning},
  1st~ed.\hskip 1em plus 0.5em minus 0.4em\relax Cambridge, MA, USA: MIT Press,
  1998.

\bibitem{deisenroth2013survey}
M.~P. Deisenroth, G.~Neumann, J.~Peters \emph{et~al.}, ``A survey on policy
  search for robotics,'' \emph{Foundations and Trends{\textregistered} in
  Robotics}, vol.~2, no. 1--2, pp. 1--142, 2013.

\bibitem{peters2006policy}
J.~Peters and S.~Schaal, ``Policy gradient methods for robotics,'' in
  \emph{Intelligent Robots and Systems, 2006 IEEE/RSJ International Conference
  on}.\hskip 1em plus 0.5em minus 0.4em\relax IEEE, 2006, pp. 2219--2225.

\bibitem{PETERS20081180}
\BIBentryALTinterwordspacing
------, ``Natural actor-critic,'' \emph{Neurocomputing}, vol.~71, no.~7, pp.
  1180 -- 1190, 2008, progress in Modeling, Theory, and Application of
  Computational Intelligenc. [Online]. Available:
  \url{http://www.sciencedirect.com/science/article/pii/S0925231208000532}
\BIBentrySTDinterwordspacing

\bibitem{schulman2015trust}
J.~Schulman, S.~Levine, P.~Abbeel, M.~Jordan, and P.~Moritz, ``Trust region
  policy optimization,'' in \emph{Proceedings of the 32nd International
  Conference on Machine Learning (ICML-15)}, 2015, pp. 1889--1897.

\bibitem{lee2001stock}
J.~W. Lee, ``Stock price prediction using reinforcement learning,'' in
  \emph{Industrial Electronics, 2001. Proceedings. ISIE 2001. IEEE
  International Symposium on}, vol.~1.\hskip 1em plus 0.5em minus 0.4em\relax
  IEEE, 2001, pp. 690--695.

\bibitem{moody2001learning}
J.~Moody and M.~Saffell, ``Learning to trade via direct reinforcement,''
  \emph{IEEE transactions on neural Networks}, vol.~12, no.~4, pp. 875--889,
  2001.

\bibitem{deng2017deep}
Y.~Deng, F.~Bao, Y.~Kong, Z.~Ren, and Q.~Dai, ``Deep direct reinforcement
  learning for financial signal representation and trading,'' \emph{IEEE
  transactions on neural networks and learning systems}, vol.~28, no.~3, pp.
  653--664, 2017.

\bibitem{mnih2013playing}
V.~Mnih, K.~Kavukcuoglu, D.~Silver, A.~Graves, I.~Antonoglou, D.~Wierstra, and
  M.~Riedmiller, ``Playing atari with deep reinforcement learning,''
  \emph{arXiv preprint arXiv:1312.5602}, 2013.

\bibitem{mnih2015human}
V.~Mnih, K.~Kavukcuoglu, D.~Silver, A.~A. Rusu, J.~Veness, M.~G. Bellemare,
  A.~Graves, M.~Riedmiller, A.~K. Fidjeland, G.~Ostrovski \emph{et~al.},
  ``Human-level control through deep reinforcement learning,'' \emph{Nature},
  vol. 518, no. 7540, pp. 529--533, 2015.

\bibitem{silver2016mastering}
D.~Silver, A.~Huang, C.~J. Maddison, A.~Guez, L.~Sifre, G.~Van Den~Driessche,
  J.~Schrittwieser, I.~Antonoglou, V.~Panneershelvam, M.~Lanctot \emph{et~al.},
  ``Mastering the game of go with deep neural networks and tree search,''
  \emph{Nature}, vol. 529, no. 7587, pp. 484--489, 2016.

\bibitem{tesauro2012simulation}
G.~Tesauro, D.~Gondek, J.~Lenchner, J.~Fan, and J.~M. Prager, ``Simulation,
  learning, and optimization techniques in watson's game strategies,''
  \emph{IBM Journal of Research and Development}, vol.~56, no. 3.4, pp. 16--1,
  2012.

\bibitem{tesauro1995td}
G.~Tesauro, ``Td-gammon: A self-teaching backgammon program,'' in
  \emph{Applications of Neural Networks}.\hskip 1em plus 0.5em minus
  0.4em\relax Springer, 1995, pp. 267--285.

\bibitem{ernst2004power}
D.~Ernst, M.~Glavic, and L.~Wehenkel, ``Power systems stability control:
  Reinforcement learning framework,'' \emph{IEEE transactions on power
  systems}, vol.~19, no.~1, pp. 427--435, 2004.

\bibitem{kober2012reinforcement}
J.~Kober and J.~Peters, ``Reinforcement learning in robotics: A survey,'' in
  \emph{Reinforcement Learning}.\hskip 1em plus 0.5em minus 0.4em\relax
  Springer, 2012, pp. 579--610.

\bibitem{sutton2000policy}
R.~S. Sutton, D.~A. McAllester, S.~P. Singh, and Y.~Mansour, ``Policy gradient
  methods for reinforcement learning with function approximation,'' in
  \emph{Advances in neural information processing systems}, 2000, pp.
  1057--1063.

\bibitem{dayan1997using}
P.~Dayan and G.~E. Hinton, ``Using expectation-maximization for reinforcement
  learning,'' \emph{Neural Computation}, vol.~9, no.~2, pp. 271--278, 1997.

\bibitem{moriarty1999evolutionary}
D.~E. Moriarty, A.~C. Schultz, and J.~J. Grefenstette, ``Evolutionary
  algorithms for reinforcement learning,'' 1999.

\bibitem{konda2000actor}
V.~R. Konda and J.~N. Tsitsiklis, ``Actor-critic algorithms,'' in
  \emph{Advances in neural information processing systems}, 2000, pp.
  1008--1014.

\bibitem{sukhbaatar2015mazebase}
S.~Sukhbaatar, A.~Szlam, G.~Synnaeve, S.~Chintala, and R.~Fergus, ``Mazebase: A
  sandbox for learning from games,'' \emph{arXiv preprint arXiv:1511.07401},
  2015.

\bibitem{sutton1999between}
R.~S. Sutton, D.~Precup, and S.~Singh, ``Between mdps and semi-mdps: A
  framework for temporal abstraction in reinforcement learning,''
  \emph{Artificial intelligence}, vol. 112, no. 1-2, pp. 181--211, 1999.

\bibitem{parr1998hierarchical}
R.~E. Parr, \emph{Hierarchical control and learning for Markov decision
  processes}.\hskip 1em plus 0.5em minus 0.4em\relax University of California,
  Berkeley Berkeley, CA, 1998.

\bibitem{parr1998reinforcement}
R.~Parr and S.~J. Russell, ``Reinforcement learning with hierarchies of
  machines,'' in \emph{Advances in neural information processing systems},
  1998, pp. 1043--1049.

\bibitem{dietterich2000hierarchical}
T.~G. Dietterich, ``Hierarchical reinforcement learning with the maxq value
  function decomposition,'' \emph{Journal of Artificial Intelligence Research},
  vol.~13, no.~1, pp. 227--303, 2000.

\bibitem{VienNLC14}
N.~A. Vien, H.~Q. Ngo, S.~Lee, and T.~Chung, ``Approximate planning for
  bayesian hierarchical reinforcement learning,'' \emph{Appl. Intell.},
  vol.~41, no.~3, pp. 808--819, 2014.

\bibitem{VienT15}
N.~A. Vien and M.~Toussaint, ``Hierarchical monte-carlo planning,'' in
  \emph{Proceedings of the Twenty-Ninth {AAAI} Conference on Artificial
  Intelligence, January 25-30, 2015, Austin, Texas, {USA.}}, 2015, pp.
  3613--3619.

\bibitem{VienLC16}
N.~A. Vien, S.~Lee, and T.~Chung, ``Bayes-adaptive hierarchical mdps,''
  \emph{Appl. Intell.}, vol.~45, no.~1, pp. 112--126, 2016.

\bibitem{barto2003recent}
A.~G. Barto and S.~Mahadevan, ``Recent advances in hierarchical reinforcement
  learning,'' \emph{Discrete Event Dynamic Systems}, vol.~13, no.~4, pp.
  341--379, 2003.

\bibitem{bellemare2016unifying}
M.~Bellemare, S.~Srinivasan, G.~Ostrovski, T.~Schaul, D.~Saxton, and R.~Munos,
  ``Unifying count-based exploration and intrinsic motivation,'' in
  \emph{Advances in Neural Information Processing Systems}, 2016, pp.
  1471--1479.

\bibitem{bacon2017option}
P.-L. Bacon, J.~Harb, and D.~Precup, ``The option-critic architecture.'' 2017.

\bibitem{fox2017multi}
R.~Fox, S.~Krishnan, I.~Stoica, and K.~Goldberg, ``Multi-level discovery of
  deep options,'' \emph{arXiv preprint arXiv:1703.08294}, 2017.

\bibitem{lee2017micro}
S.~Lee, S.-W. Lee, J.~Choi, D.-H. Kwak, and B.-T. Zhang, ``Micro-objective
  learning: Accelerating deep reinforcement learning through the discovery of
  continuous subgoals,'' \emph{arXiv preprint arXiv:1703.03933}, 2017.

\bibitem{vezhnevets2017feudal}
A.~S. Vezhnevets, S.~Osindero, T.~Schaul, N.~Heess, M.~Jaderberg, D.~Silver,
  and K.~Kavukcuoglu, ``Feudal networks for hierarchical reinforcement
  learning,'' \emph{arXiv preprint arXiv:1703.01161}, 2017.

\bibitem{durugkar2016deep}
I.~P. Durugkar, C.~Rosenbaum, S.~Dernbach, and S.~Mahadevan, ``Deep
  reinforcement learning with macro-actions,'' \emph{arXiv preprint
  arXiv:1606.04615}, 2016.

\bibitem{arulkumaran2016classifying}
K.~Arulkumaran, N.~Dilokthanakul, M.~Shanahan, and A.~A. Bharath, ``Classifying
  options for deep reinforcement learning,'' \emph{arXiv preprint
  arXiv:1604.08153}, 2016.

\bibitem{dayan1993feudal}
P.~Dayan and G.~E. Hinton, ``Feudal reinforcement learning,'' in \emph{Advances
  in neural information processing systems}, 1993, pp. 271--278.

\bibitem{kulkarni2016hierarchical}
T.~D. Kulkarni, K.~Narasimhan, A.~Saeedi, and J.~Tenenbaum, ``Hierarchical deep
  reinforcement learning: Integrating temporal abstraction and intrinsic
  motivation,'' in \emph{Advances in Neural Information Processing Systems},
  2016, pp. 3675--3683.

\bibitem{chiu2011subgoal}
C.-C. Chiu and V.-W. Soo, ``Subgoal identifications in reinforcement learning:
  A survey,'' in \emph{Advances in Reinforcement Learning}.\hskip 1em plus
  0.5em minus 0.4em\relax InTech, 2011.

\bibitem{stolle2004automated}
M.~Stolle, ``Automated discovery of options in reinforcement learning,'' Ph.D.
  dissertation, McGill University, 2004.

\bibitem{murphy2000survey}
K.~P. Murphy, ``A survey of pomdp solution techniques,'' \emph{environment},
  vol.~2, p.~X3, 2000.

\bibitem{hausknecht2015deep}
M.~Hausknecht and P.~Stone, ``Deep recurrent q-learning for partially
  observable mdps,'' 2015.

\bibitem{egorov2015deep}
M.~Egorov, ``Deep reinforcement learning with pomdps,'' 2015.

\bibitem{white1976procedures}
C.~C. White, ``Procedures for the solution of a finite-horizon, partially
  observed, semi-markov optimization problem,'' \emph{Operations Research},
  vol.~24, no.~2, pp. 348--358, 1976.

\bibitem{kaelbling1998planning}
L.~P. Kaelbling, M.~L. Littman, and A.~R. Cassandra, ``Planning and acting in
  partially observable stochastic domains,'' \emph{Artificial intelligence},
  vol. 101, no.~1, pp. 99--134, 1998.

\bibitem{Goodfellow-et-al-2016}
I.~Goodfellow, Y.~Bengio, and A.~Courville, \emph{Deep Learning}.\hskip 1em
  plus 0.5em minus 0.4em\relax MIT Press, 2016.

\bibitem{van2016deep}
H.~Van~Hasselt, A.~Guez, and D.~Silver, ``Deep reinforcement learning with
  double q-learning.'' 2016.

\bibitem{wang2015dueling}
Z.~Wang, T.~Schaul, M.~Hessel, H.~Van~Hasselt, M.~Lanctot, and N.~De~Freitas,
  ``Dueling network architectures for deep reinforcement learning,''
  \emph{arXiv preprint arXiv:1511.06581}, 2015.

\bibitem{schaul2015prioritized}
T.~Schaul, J.~Quan, I.~Antonoglou, and D.~Silver, ``Prioritized experience
  replay,'' \emph{arXiv preprint arXiv:1511.05952}, 2015.

\bibitem{gu2016continuous}
S.~Gu, T.~Lillicrap, I.~Sutskever, and S.~Levine, ``Continuous deep q-learning
  with model-based acceleration,'' in \emph{International Conference on Machine
  Learning}, 2016, pp. 2829--2838.

\bibitem{lillicrap2015continuous}
T.~P. Lillicrap, J.~J. Hunt, A.~Pritzel, N.~Heess, T.~Erez, Y.~Tassa,
  D.~Silver, and D.~Wierstra, ``Continuous control with deep reinforcement
  learning,'' \emph{arXiv preprint arXiv:1509.02971}, 2015.

\bibitem{lample2017playing}
G.~Lample and D.~S. Chaplot, ``Playing fps games with deep reinforcement
  learning.'' 2017.

\bibitem{diuk2008object}
C.~Diuk, A.~Cohen, and M.~L. Littman, ``An object-oriented representation for
  efficient reinforcement learning,'' in \emph{Proceedings of the 25th
  international conference on Machine learning}.\hskip 1em plus 0.5em minus
  0.4em\relax ACM, 2008, pp. 240--247.

\bibitem{ryan2000intrinsic}
R.~M. Ryan and E.~L. Deci, ``Intrinsic and extrinsic motivations: Classic
  definitions and new directions,'' \emph{Contemporary educational psychology},
  vol.~25, no.~1, pp. 54--67, 2000.

\bibitem{stout2005intrinsically}
A.~Stout, G.~D. Konidaris, and A.~G. Barto, ``Intrinsically motivated
  reinforcement learning: A promising framework for developmental robot
  learning,'' MASSACHUSETTS UNIV AMHERST DEPT OF COMPUTER SCIENCE, Tech. Rep.,
  2005.

\bibitem{barto2004intrinsically}
A.~G. Barto, ``Intrinsically motivated learning of hierarchical collections of
  skills,'' 2004, pp. 112--119.

\bibitem{singh2010intrinsically}
S.~Singh, R.~L. Lewis, A.~G. Barto, and J.~Sorg, ``Intrinsically motivated
  reinforcement learning: An evolutionary perspective,'' \emph{IEEE
  Transactions on Autonomous Mental Development}, vol.~2, no.~2, pp. 70--82,
  2010.

\bibitem{frank2014curiosity}
M.~Frank, J.~Leitner, M.~Stollenga, A.~F{\"o}rster, and J.~Schmidhuber,
  ``Curiosity driven reinforcement learning for motion planning on humanoids,''
  \emph{Frontiers in neurorobotics}, vol.~7, p.~25, 2014.

\bibitem{mohamed2015variational}
S.~Mohamed and D.~J. Rezende, ``Variational information maximisation for
  intrinsically motivated reinforcement learning,'' in \emph{Advances in neural
  information processing systems}, 2015, pp. 2125--2133.

\bibitem{schmidhuber2010formal}
J.~Schmidhuber, ``Formal theory of creativity, fun, and intrinsic motivation
  (1990--2010),'' \emph{IEEE Transactions on Autonomous Mental Development},
  vol.~2, no.~3, pp. 230--247, 2010.

\bibitem{tensorflow2015-whitepaper}
M.~Abadi, A.~Agarwal, P.~Barham, E.~Brevdo, Z.~Chen, C.~Citro, G.~S. Corrado,
  A.~Davis, J.~Dean, M.~Devin, S.~Ghemawat, I.~Goodfellow, A.~Harp, G.~Irving,
  M.~Isard, Y.~Jia, R.~Jozefowicz, L.~Kaiser, M.~Kudlur, J.~Levenberg,
  D.~Man\'{e}, R.~Monga, S.~Moore, D.~Murray, C.~Olah, M.~Schuster, J.~Shlens,
  B.~Steiner, I.~Sutskever, K.~Talwar, P.~Tucker, V.~Vanhoucke, V.~Vasudevan,
  F.~Vi\'{e}gas, O.~Vinyals, P.~Warden, M.~Wattenberg, M.~Wicke, Y.~Yu, and
  X.~Zheng, ``{TensorFlow}: Large-scale machine learning on heterogeneous
  systems,'' 2015, software available from tensorflow.org.

\bibitem{kingma2014adam}
D.~Kingma and J.~Ba, ``Adam: A method for stochastic optimization,''
  \emph{arXiv preprint arXiv:1412.6980}, 2014.

\bibitem{openai}
G.~Brockman, V.~Cheung, L.~Pettersson, J.~Schneider, J.~Schulman, J.~Tang, and
  W.~Zaremba, ``Openai gym,'' 2016.

\end{thebibliography}




\end{document}